\documentclass[11pt]{article}
\usepackage[left=1in,right=1in,top=1in,bottom=1in]{geometry}

\usepackage[normalem]{ulem} 
\usepackage[utf8]{inputenc} 
\usepackage[T1]{fontenc}    
\usepackage{hyperref}       
\usepackage{url}            
\usepackage{booktabs}       
\usepackage{amsfonts}       
\usepackage{nicefrac}       
\usepackage{microtype}      
\usepackage{float}
\usepackage{graphicx, xcolor, ulem}
\usepackage{algorithm,algpseudocode}
\usepackage{multirow}
\usepackage{enumitem}
\usepackage{amsmath,amsfonts,latexsym,amssymb,amsbsy, amsthm, mathtools}

\usepackage{footnote}
\usepackage{thmtools, thm-restate}
\usepackage{authblk}
\usepackage{setspace}
\usepackage{caption}
\usepackage{subcaption}
\usepackage{array}
\usepackage{tabularx}

\newcolumntype{Y}[1]{>{\PreserveBackslash\centering}p{#1}}

\usepackage[symbol]{footmisc}

\newcommand \pindex[1]{\ensuremath{x[\mathrm{#1},i,t]}}
\newcommand \dindex[1]{\ensuremath{\text{data}[\mathrm{#1},i,t]}}

\newcommand \gindex[2]{\ensuremath{{#1}[\mathrm{#2},i,t]}}

\usepackage{rotating}

\usepackage{rotating, float, caption}

\algnewcommand\INPUT{\item[{\textbf{Input:}}]}
\algnewcommand\RETURN{\item[{\textbf{Return:}}]}

\allowdisplaybreaks 
\algnewcommand\OUTPUT{\item[{\textbf{Output:}}]}

\title{Improving Clinical Decision Support through Interpretable Machine Learning and Error Handling in Electronic Health Records}

\author{
Mehak Arora$^{1,2,*,\dagger}$, 
Hassan Mortagy$^{3,\dagger}$, 
Nathan Dwarshuis$^3$, 
Jeffrey Wang$^4$, 
Philip Yang$^5$, 
Andre L Holder$^5$, 
Swati Gupta$^{6,\ddagger}$, 
Rishikesan Kamaleswaran$^{1,2,\ddagger}$

$^1$Department of Electrical and Computer Engineering, Duke University, Durham, NC, USA\\
$^2$Department of Surgery, Duke University School of Medicine, Durham, NC, USA\\
$^3$Department of Industrial and Systems Engineering, Georgia Institute of Technology, Atlanta, GA, USA\\
$^4$Division of Cardiology, Emory University School of Medicine, GA, USA\\
$^5$Division of Pulmonary, Allergy, Critical Care and Sleep Medicine, Emory University School of Medicine, GA, USA\\
$^6$Sloan School of Management, Massachusetts Institute of Technology, Cambridge, MA, USA\\

$^*$Corresponding author: mehak.arora@duke.edu \\ $^\dagger$These authors contributed equally to this work. $^\ddagger$Equal senior authors. \\
This work is being published in the Journal of the American Medical Informatics Association.
}

\date{}

\begin{document}

\maketitle

\begin{abstract}
    The objective of this work is to develop an Electronic Medical Record (EMR) data processing tool that confers clinical context to Machine Learning (ML) algorithms for error handling, bias mitigation and interpretability. We present Trust-MAPS, an algorithm that translates clinical domain knowledge into high-dimensional, mixed-integer programming models that capture physiological and biological constraints on clinical measurements. EMR data is projected onto this constrained space, effectively bringing outliers to fall within a physiologically feasible range. We then compute the distance of each data point from the constrained space modeling healthy physiology to quantify deviation from the norm. These distances, termed ``trust-scores,'' are integrated into the feature space for downstream ML applications. We demonstrate the utility of Trust-MAPS by training a binary classifier for early sepsis prediction on data from the 2019 \texttt{PhysioNet} Computing in Cardiology Challenge, using the XGBoost algorithm and applying SMOTE for overcoming class-imbalance.  The Trust-MAPS framework shows desirable behavior in handling potential errors and boosting predictive performance. We achieve an AUROC of 0.91 (0.89, 0.92 : 95\% CI) for predicting sepsis 6 hours before onset - a marked 15\% improvement over a baseline model trained without Trust-MAPS. Trust-scores emerge as clinically meaningful features that not only boost predictive performance for clinical decision support tasks, but also lend interpretability to ML models. This work is the first to translate clinical domain knowledge into mathematical constraints, model cross-vital dependencies, and identify aberrations in high-dimensional medical data. Our method allows for error handling in EMR, and confers interpretability and superior predictive power to models trained for clinical decision support. 
\end{abstract}

\vspace{-10pt}
                                                                                         
\section{Background and Significance}
Complex decision-making in time-critical domains like the Intensive Care Unit (ICU) involves careful attention to information from several data sources like the patient's medical history, and their current clinical tests. Trained clinicians comprehend the limitations of each data-type in a given clinical situation and evaluate the entirety of this data, easily recognizing conflicting information. Thus, they continually navigate a multidimensional feature space grounded in physical and biological constraints. Using their domain knowledge, they can pick out outlier data points and can attribute them to errors in the data.

There is growing interest in leveraging machine learning (ML) to support clinical decision-making in tasks like ICU mortality prediction and early sepsis detection \cite{ML_CDS}. ML models are trained on retrospective patient data stored in Electronic Medical Records (EMR). However,  EMRs often contain missing or erroneous data  \cite{brundin2018secondary, phillips2009developing}. While clinicians can interpret and amend discrepancies based on medical expertise, ML pipelines do not have the clinical context to be able to make similar corrections. In this work, we address the challenge of embedding domain knowledge into machine learning pipelines to create context-aware algorithms that can automatically detect data errors, thereby reducing model bias. We also introduce a feature engineering method using domain knowledge that enhances both interpretability and predictive performance. We present this approach as a plug-and-play preprocessing tool for high-fidelity machine learning, called {\it Trust-MAPS} - \textbf{T}rustable \textbf{R}eal-time and \textbf{U}ncertainty-managed \textbf{ST}reaming \textbf{M}L \textbf{A}lgorithms using \textbf{P}rojections for \textbf{S}afety critical diagnosis. 

Error identification in large medical datasets has traditionally relied on outlier detection methods, such as inter-quartile range tests  \cite{Diabetes_outliers}, mean-variance tests  \cite{KNN_outliers}, and K-Nearest Neighbour algorithms  \cite{KNN_outliers}. Errors are typically handled using supervised learning techniques like regression models  \cite{impute_casebased, MICE, Diabetes_outliers}. However, these methods don't incorporate clinical context into the data processing pipeline. Trust-MAPS is a constrained optimization approach for detecting potential data errors and integrating clinical knowledge by translating complex biological and physiological constraints on vitals and lab values into high-dimensional mathematical constraints. The ``projection'' of data derived from the EMR onto a sub-space defined by feasible domain constraints allows us to bring outliers back into the physically-possible range of values, consequently bolstering the reliability of subsequent data-driven learning tasks. We also design a novel set of features, called \textit{trust-scores}, by computing the distance between a patient's clinical state—defined by vital signs and laboratory values—and the constrained space representing normal ranges for healthy patients. Trust-scores can offer interpretable insights to explain model predictions. We validate our framework by applying it to machine learning models trained on the \texttt{PhysioNet} Computing in Cardiology Challenge 2019  \cite{Reyna2020} for the early prediction of sepsis. While the primary focus in this study is sepsis prediction, the projection-based framework of Trust-MAPS is versatile and can generalize to other clinical decision support models. 

Recent advances in sepsis prediction leverage various machine learning (ML) techniques, including gradient-boosted trees \cite{yang2020explainable, physionet1}, deep learning architectures such as LSTM \cite{strickler2023exploring, rosnati2021mgp, gupta2024improving}, and hybrid approaches that combine these methods \cite{duan2023early}. However, challenges persist due to the susceptibility of these algorithms to biases and systemic errors present in training data, which often lead to high false alarm rates and clinician fatigue \cite{clifford2020future, gupta2024improving}. To mitigate these issues, researchers have aimed to improve ML model robustness by incorporating clinical knowledge  \cite{ma2023knowledge, JANJUA2022106638}, addressing uncertainties  \cite{yin2024sepsislab}, and improving interpretability \cite{nemati2018interpretable, strickler2023exploring, antoniadi2021current}. Some approaches for adding clinical context include knowledge graphs \cite{ma2023knowledge}, risk factor-based regularization \cite{xu2023interpretability}, expert-defined rule integration \cite{JANJUA2022106638}. Our method builds on the latter by introducing a flexibile framework that can handle high-dimensional data, accommodate range-based constraints, and capture cross-feature interactions. Methods to confer interpretability to ML algorithms include attention mechanisms for deep learning \cite{strickler2023exploring, rosnati2021mgp}, SHAP values for traditional classifiers \cite{yang2020explainable}, rule-based models with target-defined separation thresholds  \cite{Donoho2008, sheth2019univariate}, and subgroup discovery \cite{Struelens2021}. Our results show that the Trust-MAPS framework compliments these approaches well, and can be used as preprocessing step to any downstream ML pipeline.

\begin{figure}[!t]
    \includegraphics[width=\textwidth,trim={0 0.1cm 0 0},clip]{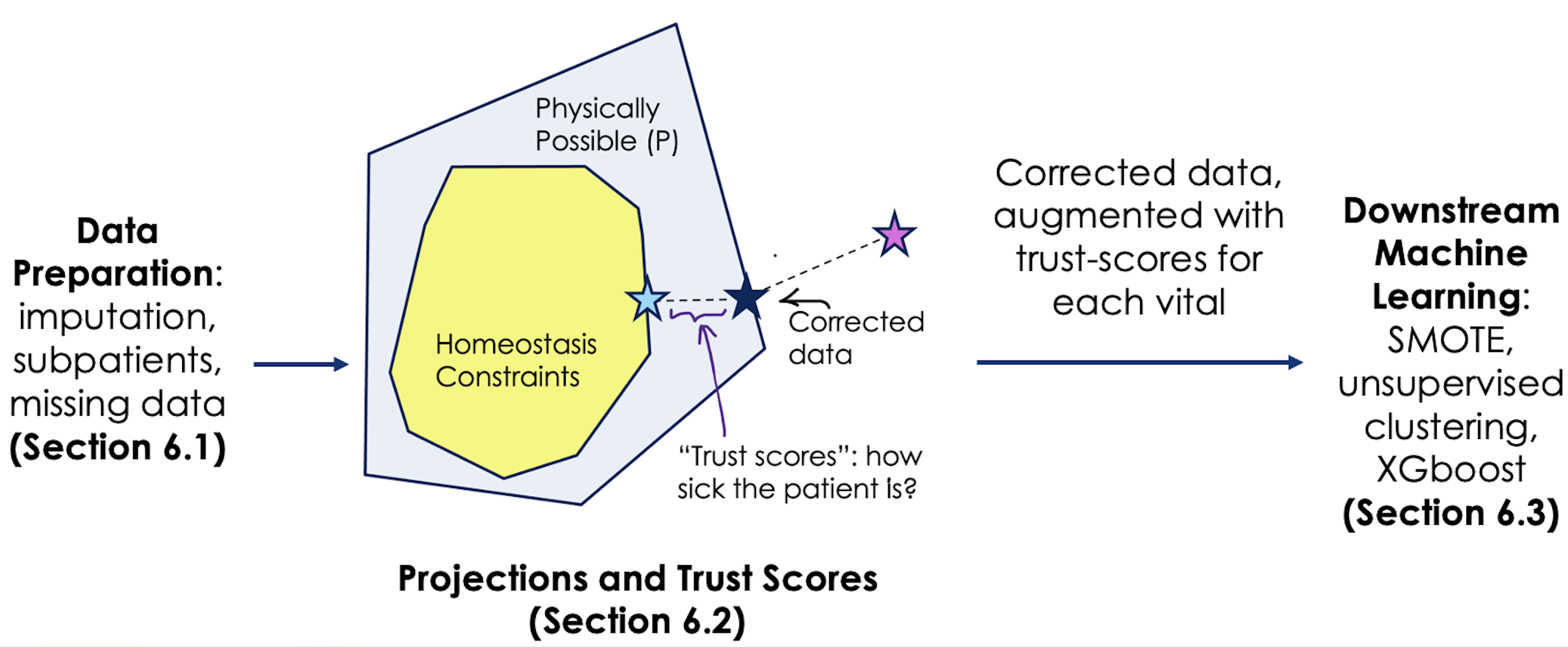}
    \caption{Trust-MAPS augment the ``corrected data'' computed using projections onto physical constraints $P$, with ``trust-scores'' which capture the distance of the corrected data from homeostasis constraints. This augmented data is then used in downstream ML for predictions, and significantly improves the performance. There are a number of steps needed to normalize the length of the stay of patients, impute missing values in the data to get the new patient data in standardized form, and then data augmentation with trust-scores for each patient vital (See Figure  \ref{fig:pipeline} in the Supplemental  \ref{sec:methods}), before using projections. The pipeline is very general, and can be used for more advanced predictive methods (e.g., those that process clinical notes).}
    \label{fig:pipeline-high}
\end{figure}

Thus, we present Trust-MAPS as a data preprocessing module with the following functionalities: (i) handling potential errors in EMR data, (ii) integrating clinical context into ML pipelines through expert-defined constraints on clinical variables, and (iii) engineering a new set of interpretable features, termed \textit{trust-scores}, that quantify a patient's feature-specific deviation from healthy physiology. 


\begin{figure}
    \centering
    
    \begin{minipage}[t]{0.5\textwidth}
    \centering
    \includegraphics[width=\linewidth]{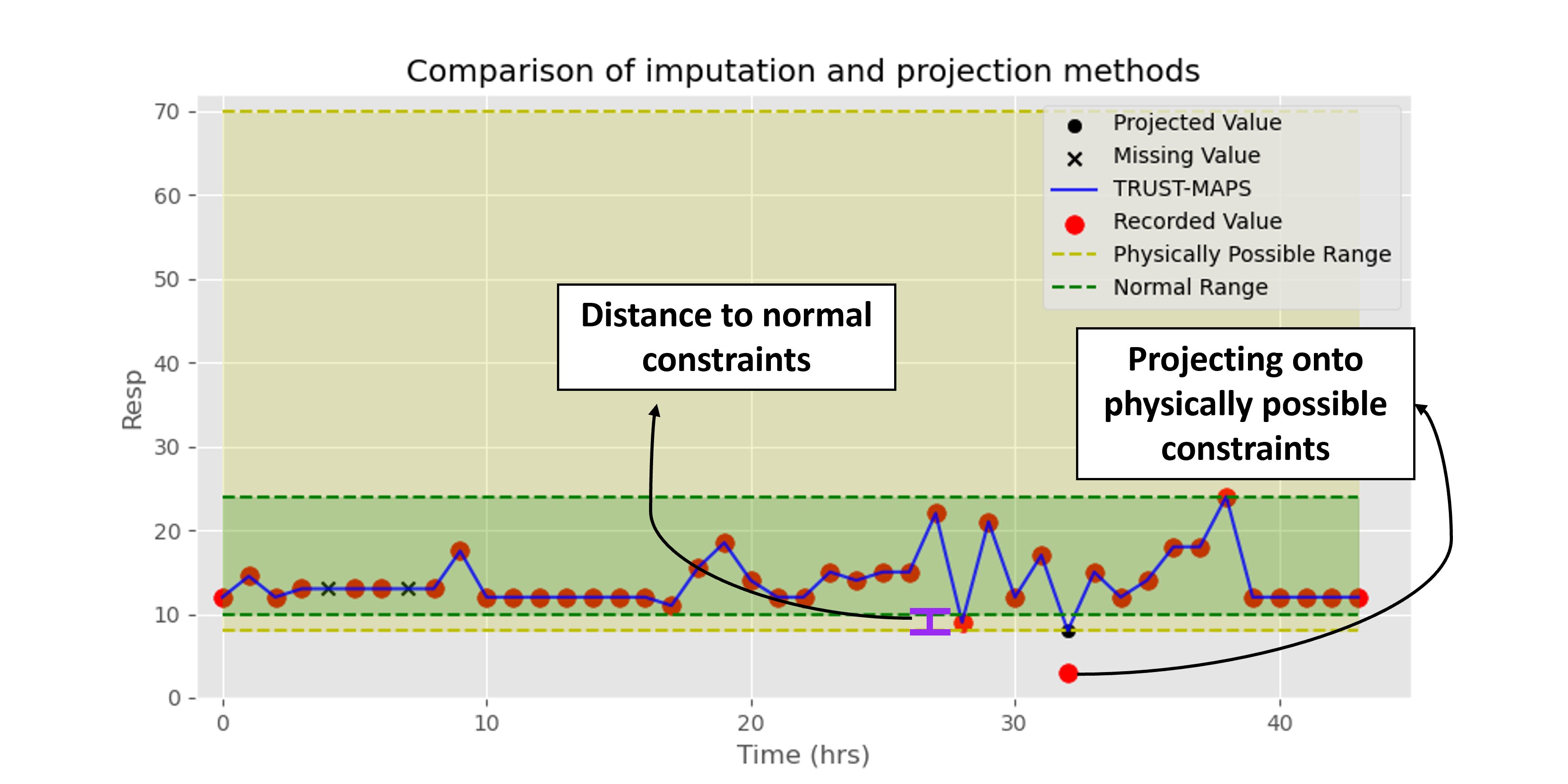}
    \subcaption{Example of Trust-MAPS on Respiratory Rate}
    \label{fio2}
    \end{minipage}\hfill
    \begin{minipage}[t]{0.5\textwidth}
    \centering
    \includegraphics[width=\linewidth]{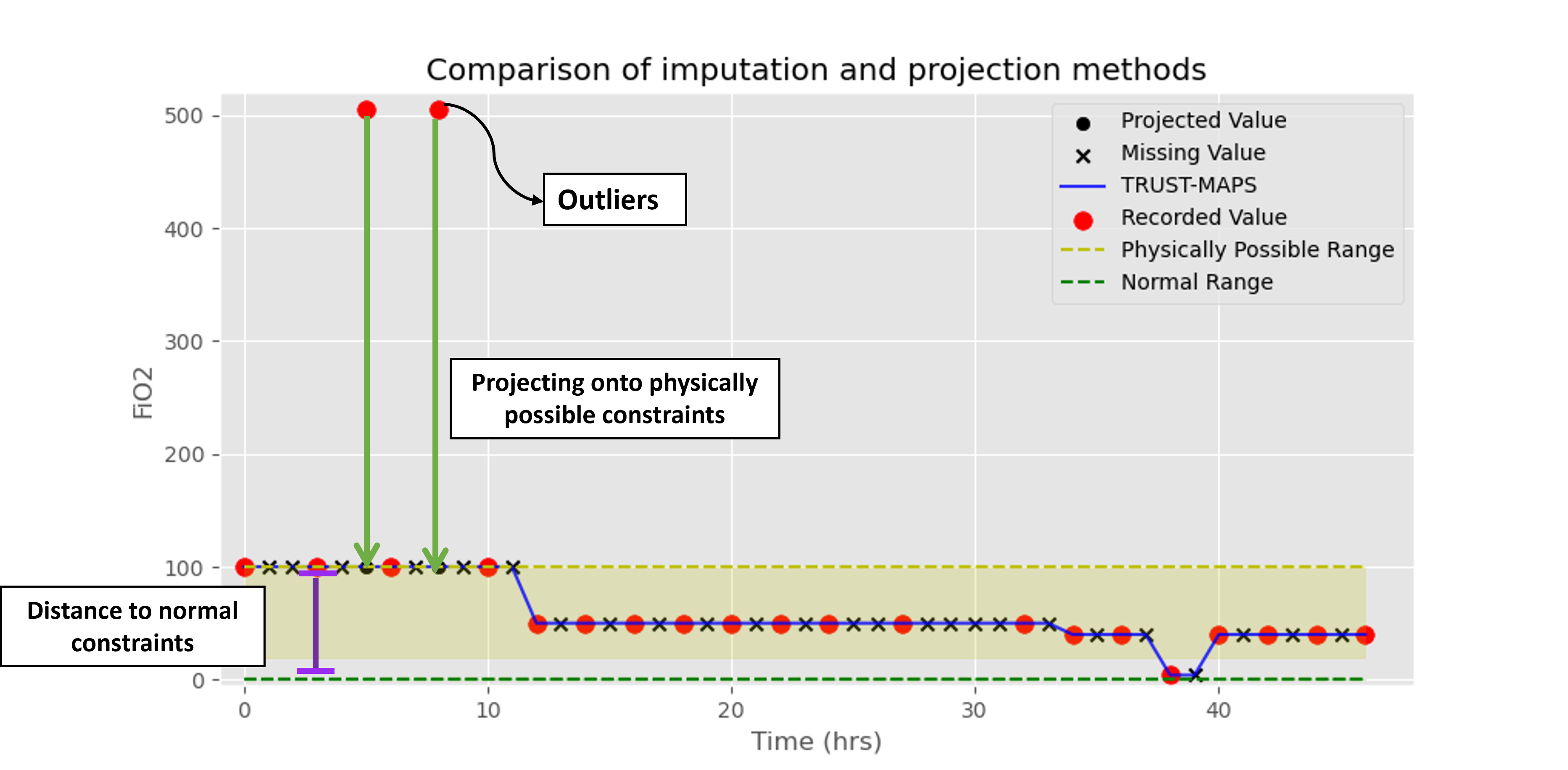}
    \subcaption{Example of Trust-MAPS on FiO2}
    \label{resp}
    \end{minipage}

    \caption{ A visualization of the Trust-MAPS processing on clinical data. In the first step, outlier data points that lie outside the physically possible constraints are projected to lie within the constraints. This is to prevent the machine learning model from learning incorrect patterns from erroneous data and mitigate potential bias in the downstream learning task. In the second step, the distance to the normal constraints is calculated to provide a cohesive number that quantifies how deviated the patient is, in any particular interval of time (\textit{sub-patient}), from normal physiology.}
    \label{fig:projections}
\end{figure}

\section{Methods} \label{sec: Trust-MAPS}

A \textit{projection} is a fundamental operation that arises in many different fields, ranging from architecture to machine learning (e.g.,  \cite{duchi2008efficient}). The projection of a given point $x \in \mathbb{R}^n$ in high-dimensional space to some set $S \subseteq \mathbb{R}^n$ is the ``closest'' point in the set $S$ to $x$. For example, a point in 3-dimensional space (e.g., a ball in the air) will project onto its shadow on the floor. 

The key idea of Trust-MAPS is to model prior knowledge of feasible data values as high-dimensional mathematical constraints. These constraints could be upper and lower bounds on feature values, physically-possible rate of change in recorded values, or conditional interactions between variables. We define the following two sets of clinically interpretable constraints:

\begin{itemize}
    \item \textit{Physical Constraints} (set $P$): These constraints define biologically possible bounds on values and rates of change of clinical data of critically ill ICU patients, as well as dependencies between various clinical measurements. For example, the physical range for heart rate is 30-200 beats per minute. These constraints also capture the rate of change of various vitals and lab values, as well as dependencies across different values, e.g., if {\it bicarbonate} is less than 10, then {\it base excess} must be at most 0 (detailed in Section  \ref{sec:phys proj}).
    
    \item \textit{Normal Constraints} or \textit{Homeostasis Constraints} (set $N$): These are constraints on the expected range of values of a healthy patient's data, for example, the normal range for heart rate is 60-90 beats per minute, along with rate constraints and dependencies across lab values and vitals (detailed in Section  \ref{sec:norm proj}).    
\end{itemize}

We are given observed, temporal patient data, $\texttt{data[v, t]}$ from the EMR, where $v \in V$, the set of all vitals and lab values in consideration. The projection, $x[v,t]$ of $\texttt{data[v,t]}$ onto the set of constraints $P$ is a way to determine which values in \texttt{data[v,t]} are erroneous, and to bring these values into a clinically feasibility range. The goal is to prevent the machine learning model from learning incorrect patterns from erroneous values, and to avoid scenarios in which single, abnormal events significantly impact decision-making. On the other hand,the distance of the projected data $x[v,t]$ to constraints capturing normal physiology (set $N$) are informative of how sick the patient is, and enrich the feature space for downstream ML. A high-level depiction of this pipeline can be seen in Figure  \ref{fig:projections}.

The length of temporal patient data varies with the duration of ICU stay. To apply Trust-MAPS, it is necessary to create a data point of fixed dimensions. One way would be to aggregate feature recordings every hour and treat each hour as a separate data-point. However, temporal trends, which can be informative in clinical decision-making, are lost. For this study, we divide patient data into fixed time intervals, referred to as \textit{sub-patients}, with each \textit{sub-patient} treated as an individual data point. A \textit{sub-patient} provides a snapshot-in-time of the patient's overall ICU stay. For a fixed time interval $n$ and number of clinical variables $f$, we generate snapshots of size $n \times f$, unroll each clinical feature, and create one-dimensional sub-patients of length $n*f$. A detailed rationale and methodology for creating \textit{sub-patients} is provided in Supplemental Section  \ref{sec: stand}. For the results in this paper, we use 6-hour intervals with a 3-hour overlap, and include similar plots for other intervals and overlaps in the Supplemental Materials (Section  \ref{app:different_time_window_sec}). Data is pre-processed by imputing missing values (Supplemental Section  \ref{sec:prep}), then applying min-max normalization (Supplemental Section  \ref{sec:norm}). 

The Trust-MAPS pipeline is a two-step process. First, we project patient data onto set $P$,  as described in detail in Supplemental Section  \ref{sec:phys proj}. The resulting data is considered free of outliers and erroneous values. Next, we calculate the euclidean distance of the corrected data to its projection on the decision set $N$. For a given variable $v$ in a \textit{sub-patient} of time-interval $n$, if $v^P = [{v^P}_1, {v^P}_2, \dots, {v^P}_n]$ represents the vector of values of $v$ recorded during that \textit{sub-patient} interval and projected onto $P$, and $v^N = [{v^N}_1, {v^N}_2, \dots, {v^N}_n]$ represents the projection of $v^P$ onto $N$, then the euclidean distance $||v^P - v^N||$ is what we call the distance-to-normal or ``trust-score'' for $v$ in that \textit{sub-patient}. Trust-scores give us a trustworthy and interpretable measure of how sick a patient might be. They reduce a sequence of $n$ values recorded over a period of time to a single number that quantifies the deviation from healthy physiology, with respect to each clinical variable (Equation  \ref{NormDistance}). We subsequently use trust-scores as input features for training a machine learning model for prediction of sepsis. This process is described in Supplemental Section  \ref{sec:norm proj}, see Figure  \ref{fig:pipeline-high}. The complete list of upper and lower bounds on clinical data used to define $N$ and $P$, and the rate constraints, can be found in Supplemental Section  \ref{app:bound}, and the detailed formulation of our {\it if-then-else} constraints can be found in Supplemental Section  \ref{app:proj}. These clinical constraints were formulated by a panel of three board-certified physicians, following a consensus achieved through collaborative discussions.

\section{Results}\label{sec:results}
\subsection{Data Description and Summary of Clinical Characteristics} \label{sec: clinic}
\begin{figure}[H]
\centering
\begin{subfigure}{\textwidth}
  \centering
  \includegraphics[scale = 0.8, width=\linewidth]{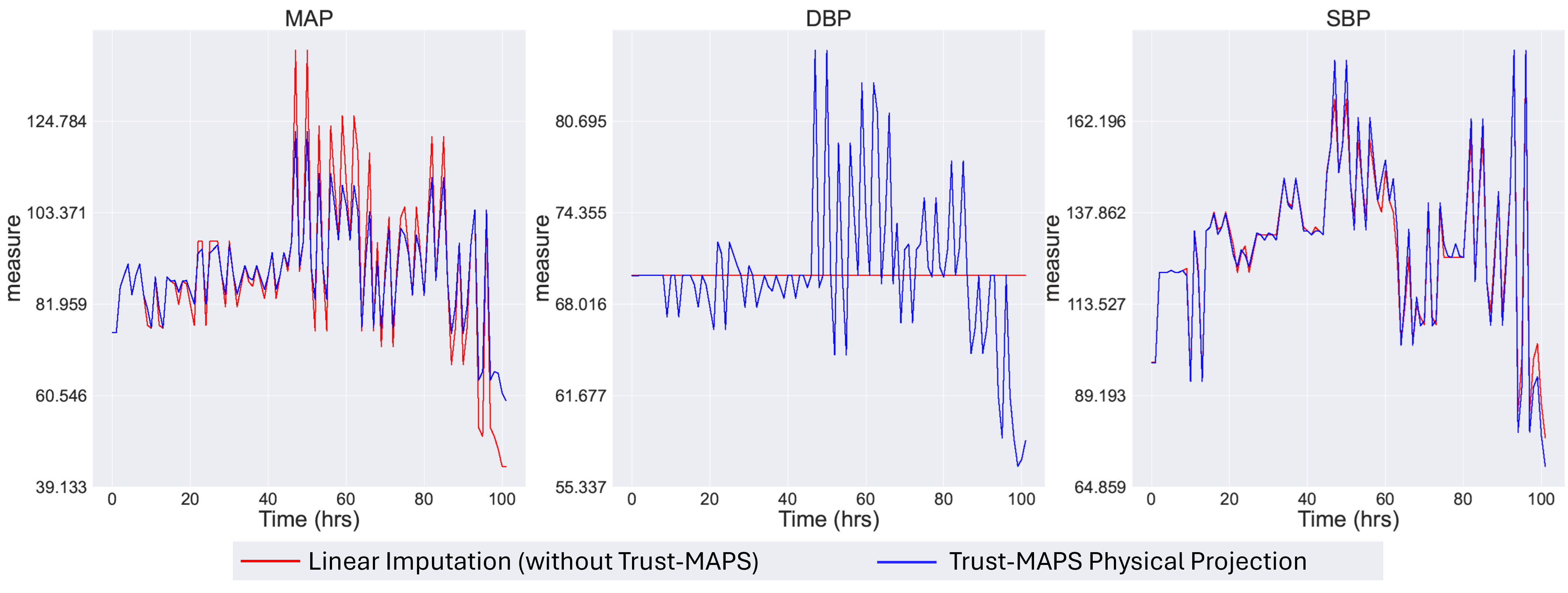}
  \caption{Non-linear Interactions between MAP, DBP, SBP}
  \label{fig:psub1}
\end{subfigure}%

\begin{subfigure}{0.49\textwidth}
  \centering
  \includegraphics[width=\linewidth]{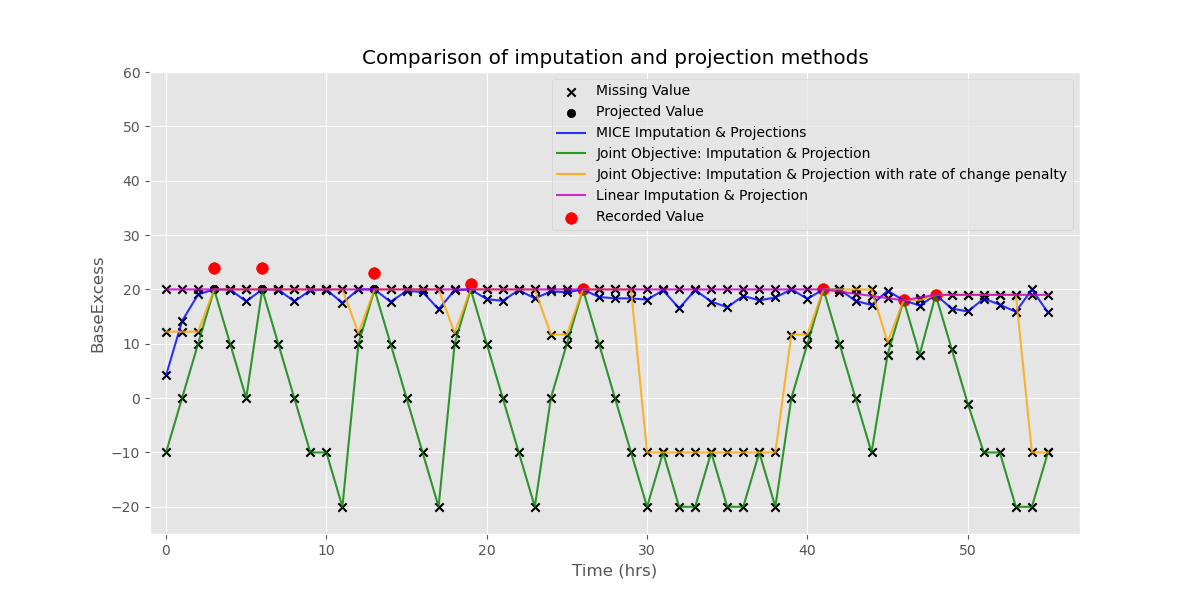}
  \caption{An Example of Imputation of Base Excess}
  \label{fig:psub2}
\end{subfigure}
\hfill
\begin{subfigure}{0.49\textwidth}
  \centering
  \includegraphics[width=\linewidth]{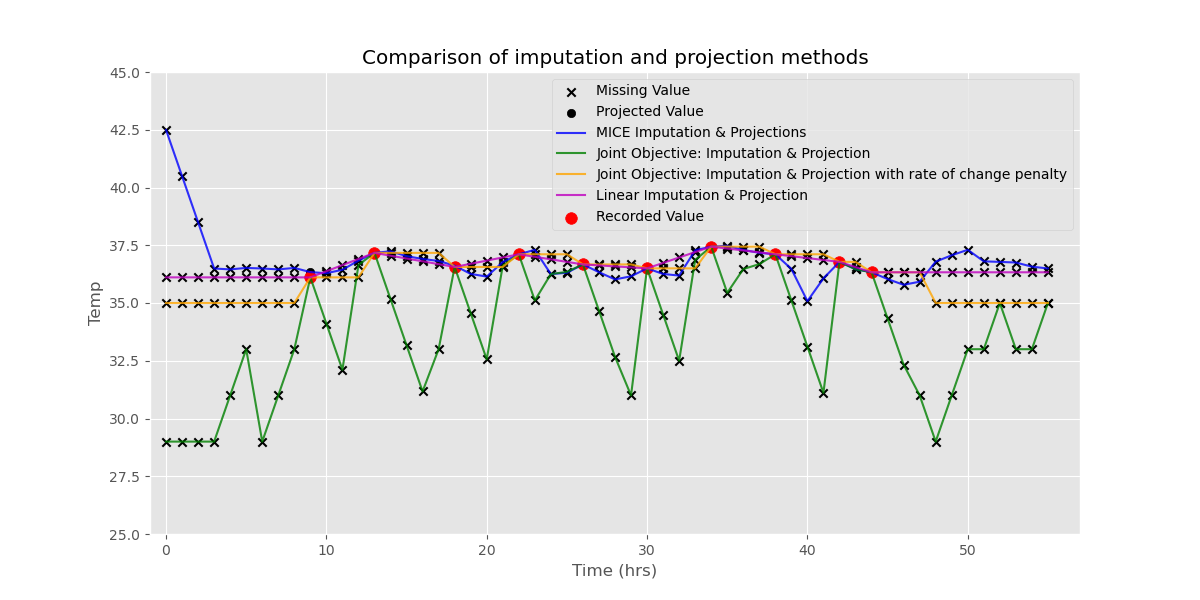}
  \caption{An Example of Imputation of Temperature}
  \label{fig:psub3}
\end{subfigure}

\caption{\small An illustrative example of data correction by projection on the Physical Constraints Set $P$ demonstrated through (a) the constraint on the relationship between MAP, DBP and SBP, where the imputed values of DBP default to the average value but the projected values satisfy the interactions between these vitals given by constraint \texttt{\#4} in Supplemental Section  \ref{sec:phys proj}, (b) A comparison of imputation of Base Excess by methods described in the Supplemental Section  \ref{app:alt_imputation}. The Trust-MAPS approach is labeled ``Linear Imputation and Projections.'' (c) A comparison of imputation of temperature. }
\label{fig:proj behav}
\end{figure}

We assess the performance of Trust-MAPS on a publicly-available dataset released as part of the \texttt{PhysioNet} Computing in Cardiology Challenge 2019  \cite{Reyna2020}\footnote{The data can be accessed at \url{https://physionet.org/content/challenge-2019/1.0.0/}}. The data consists of six demographic features, eight vital signs, and 26 laboratory variables aggregated into hourly bins and recorded over a patient's stay in the ICU. Thirty clinical variables from the \texttt{PhysioNet} dataset were used for building our model. These are enlisted along with their expert-defined physical and normal range constraints in {Table  \ref{tab:ranges} of the Supplemental  \ref{app:bound}. We compute the SIRS (Systemic Inflammatory Response System)  \cite{Bone1992} and SOFA (Sepsis Related Organ Failure Assessment)  \cite{vincent1996sofa} score for benchmarking our ML algorithm\footnote{We also use SOFA and SIRS as predictors in our ML model to improve prediction power and interpretability.}. Each patient record has an associated time-dependent ``Sepsis Label'' column which is set to 1 six hours before the onset of sepsis. 
The time at which the patient develops sepsis is decided according to the Sepsis-3 definition  \cite{seymour2016assessment}, which is given in {Supplemental  \ref{app:sepsis}}. The data consisted of 40,336 patients, out of which  2,932 patients (7.2\%) developed sepsis. After standardizing to fixed ``sub-patient'' time windows, we obtained a total of 463,693 sub-patients, out of which 9,646 (2\%) correspond to sepsis patients.   Table  \ref{tab:char} in the Supplemental  \ref{app:bound} details the complete descriptive statistics of the dataset.

\subsection{Data Correction by Projection onto Physical Constraints} \label{sec: physical proj results}

Projecting data onto $P$, defined as a ''physical projection'', can bring outlier data points to a value within the physically-possible constraints. The constraints we define account for their minimum and maximum values, maximum rate of change values, and interactions between variables. Figure  \ref{fig:proj behav} illustrates the physical projection step demonstrated through constraint on the relationship between Mean Arterial Pressure (MAP), Systolic Blood Pressure (SBP), and Diastolic Blood Pressure (DBP) ( \ref{sec:phys proj}  Constraint \texttt{\#4}). In this case, there was no recorded value of DBP in the EMR. Trust-MAPS first imputes DBP to a constant value equaling the mean of the ''normal'' range of values, i.e., 70. However, the physical projections step captures the interaction between DBP, SBP, and MAP and autocorrects DBP accordingly. We refer the reader to Figures  \ref{fig:proj behav 3}- \ref{fig:resp} in Supplemental Section  \ref{app:comp} for more interesting examples of data correction related to phosphate, respiration, base excess, bicarbonate, and lactate.}

We compare Trust-MAPS with alternate methods of imputation, as detailed in the Supplemental Materials Section  \ref{app:alt_imputation}. Particularly, we test Trust-MAPS against MICE  \cite{MICE}, a commonly used imputation technique for healthcare data, along with alternate formulation of the optimization objective used for projections. The behavior of different imputation strategies is displayed in Figure  \ref{fig:psub2}, with more examples in the Supplemental Figure  \ref{fig:alternate_imp}. We can see that the linear imputation and projections approach used by Trust-MAPS shows desirable behaviors in correcting outlier values and faithfully imputing missing values. 

A caveat to note is that the distance of a data point from the constraint set $P$ does not always reflect the true extent of error in the variable. For example, a misplaced decimal point can result in a large distance. However, any non-zero distance still signals potential data errors and informs confidence in subsequent model predictions. \footnote{Most laboratory values are transmitted automatically from lab machines to the EMR. Thus, detected errors might indicate issues in sample collection or laboratory procedures, which we do not address in this work.}

\subsection{Trust-MAPs Case Study: Sepsis Prediction} \label{sec: ml results}

\subsubsection{Superior Classifier Performance for Sepsis Prediction}
 The machine-learning pipeline used in this study was designed to combat the severe class imbalance observed in the data and to maximize precision (Supplemental Section  \ref{sec:ML}). Given the highly heterogeneous nature of patient data, we follow the Cluster-then-Predict approach  \cite{ClusterPredict} to maximize performance. We perform unsupervised $k$-means clustering with $25$ clusters on the training data and train a different XGBoost \cite{chen2016xgboost} on each cluster. Details regarding the choice of the number of clusters and methods of analysis are described in Supplemental Section  \ref{sec:ML}, and results are presented in Figure  \ref{fig:clustering} in Supplemental Section  \ref{app:comp}. 

  The baseline for comparison of classification performance for sepsis prediction is a model trained on the dataset after linear imputation, and without Trust-MAPS. We then train two separate models on data obtained after applying the first step of Trust-MAPS (physical projection), and the second step of Trust-MAPS (normal projection) respectively. As an additional comparison, we train a machine learning model after applying MICE.

\begin{figure}[h]
    \centering
    \includegraphics[width=\textwidth]{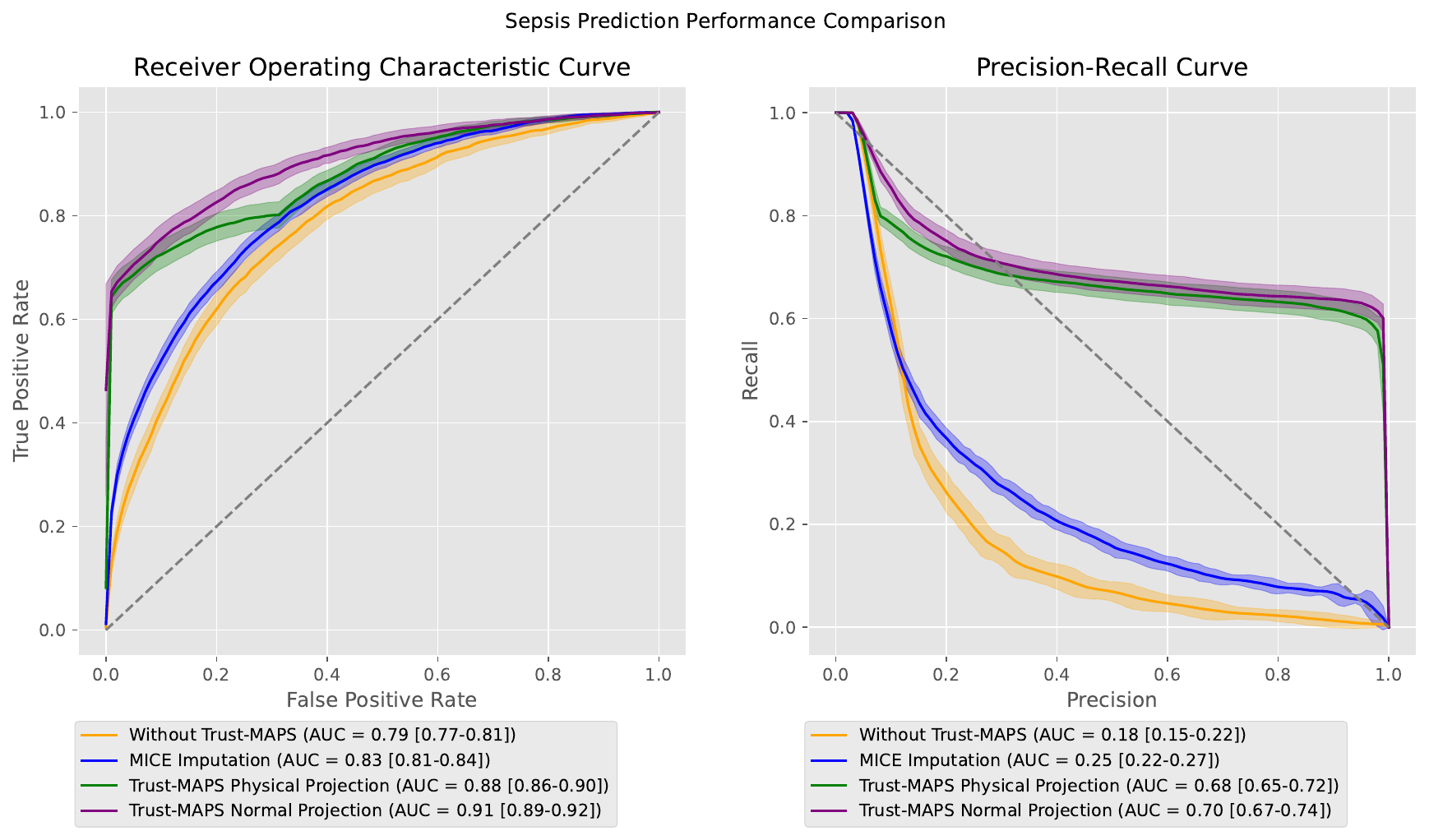}
    \caption{\textbf{Left:} A plot comparing Receiver-Operating Characteristic Curves for sepsis prediction machine learning model on the dataset processing with each step of the Trust-MAPS process, and baseline models trained without Trust-MAPS. \textbf{Right:} A plot comparing Precision-Recall Curves for sepsis prediction machine learning model on the dataset processing with each step of the Trust-MAPS process, and baseline models trained without Trust-MAPS.} 
    \label{fig:ROC}
\end{figure}

We refer the reader to Table  \ref{tab:results1} and Figure  \ref{fig:ROC} for the results comparing the performance of the data correction and imputation techniques. The model trained on data with Trust-MAPS achieves an AUC of 0.901, approximately a 15\% improvement over the baseline. The model trained with Trust-MAPS is also better calibrated, as is evident from the reliability curve and brier score results detailed in the Supplemental Materials (Section  \ref{app:calibration}.)

\begin{table}[t]
\centering
\small\addtolength{\tabcolsep}{-1pt}
\resizebox{\textwidth}{!}{
\begin{tabular}{|c|c|c|c|c|c|c|c|}
\hline 
\textbf{Dataset} & \textbf{Method} & \textbf{Sensitivity}       & \textbf{Specificity}     
& \textbf{Precision}  & \textbf{AU-ROC} & \textbf{AU-PRC} & \textbf{$f$-score} \\
\hline 
	\multirow{4}{3.1cm}{Train Set} & {Trust-MAPS Normal Projection} & 0.804 ± 0.001 & 0.999 ± 0.00002
	& 0.974 ± 0.001	& 0.961 ± 0.001 & 0.857 ± 0.001 &	0.881 ± 0.003    
	 \\
        & {Trust-MAPS Physical Projection} & 0.793 ± 0.005 &	0.999 ± 0.0003 &	0.956 ± 0.002 &	0.950 ± 0.002 &	0.846 ± 0.004 &	0.867 ± 0.003  \\ 
        & {MICE Imputation} & 0.579 ± 0.001 &	0.996 ± 0.0001 &	0.245 ± 0.002 &	0.891 ± 0.006 &	0.349 ± 0.004 &	0.344 ± 0.003  \\ 
        & {Baseline (Without Trust-MAPS)} & 0.614 ± 0.003 &	0.951 ± 0.003&	0.538 ± 0.012&	0.828 ± 0.012 &	0.199 ± 0.031&	0.552 ± 0.030  \\ 
        & {SOFA Scores} &  0.168 ± 0.012&	0.896 ± 0.001&	0.054 ± 0.021&	-	& - &	0.081 ± 0.081 \\ 
        \hline 
	\multirow{4}{3.1cm}{Test Set} & {Trust-MAPS Normal Projection} & \textbf{0.633 ± 0.018}	& \textbf{0.998 ± 0.0003}	& \textbf{0.951 ± 0.004} &	\textbf{0.907 ± 0.008} &\textbf{0.707 ± 0.017} & \textbf{0.760 ± 0.012} \\ 
    & {Trust-MAPS Physical Projection} & 0.622 ± 0.011 &	0.993 ± 0.0009 &	0.910 ± 0.005 &	0.882 ± 0.010 &	0.684 ± 0.017 &	0.739 ± 0.015  \\ 
    & {MICE Imputation} & 0.370 ± 0.002 &	0.958 ± 0.001 &	0.165 ± 0.001 &	0.833 ± 0.002 &	0.248 ± 0.001 &	0.228 ± 0.012  \\ 
     & {Baseline (Without Trust-MAPS)} &  0.445 ± 0.015 &	0.949 ± 0.003 &	0.499 ± 0.007 &	0.785 ± 0.019 &	0.176 ± 0.012 &	0.435 ± 0.014 \\ 
     & {SOFA Scores} &  0.152 ± 0.020 &	0.897 ± 0.001&	0.05 ± 0.025&	-	& - &	0.075 ± 0.074

 \\ \specialrule{.2em}{.1em}{.1em} 
\end{tabular}}
     \caption{Confidence intervals for the sensitivity, specificity, precision, AUC-ROC (Area under the Receiver Operating Characteristics Curve), AUC-PRC (Area under the Precision-Recall Curve), and $f$-score of the sepsis prediction algorithm over 50 iterations. We benchmark the performance of the machine learning learning model for sepsis prediction at each step of the Trust-MAPS pipeline against the performance of models trained without Trust-MAPS. Additionally, we compare the performance on the same machine learning pipeline on data imputed using MICE. We also benchmark against using the SOFA score as an indicator of sepsis. For this purpose, a SOFA score greater than or equal to 2 indicates that the patient has developed sepsis  \cite{vincent1996sofa}.}
    \label{tab:results1}
\end{table}

\begin{figure}
\centering
\begin{subfigure}{.5\textwidth}
  \centering
  \includegraphics[width=\linewidth]{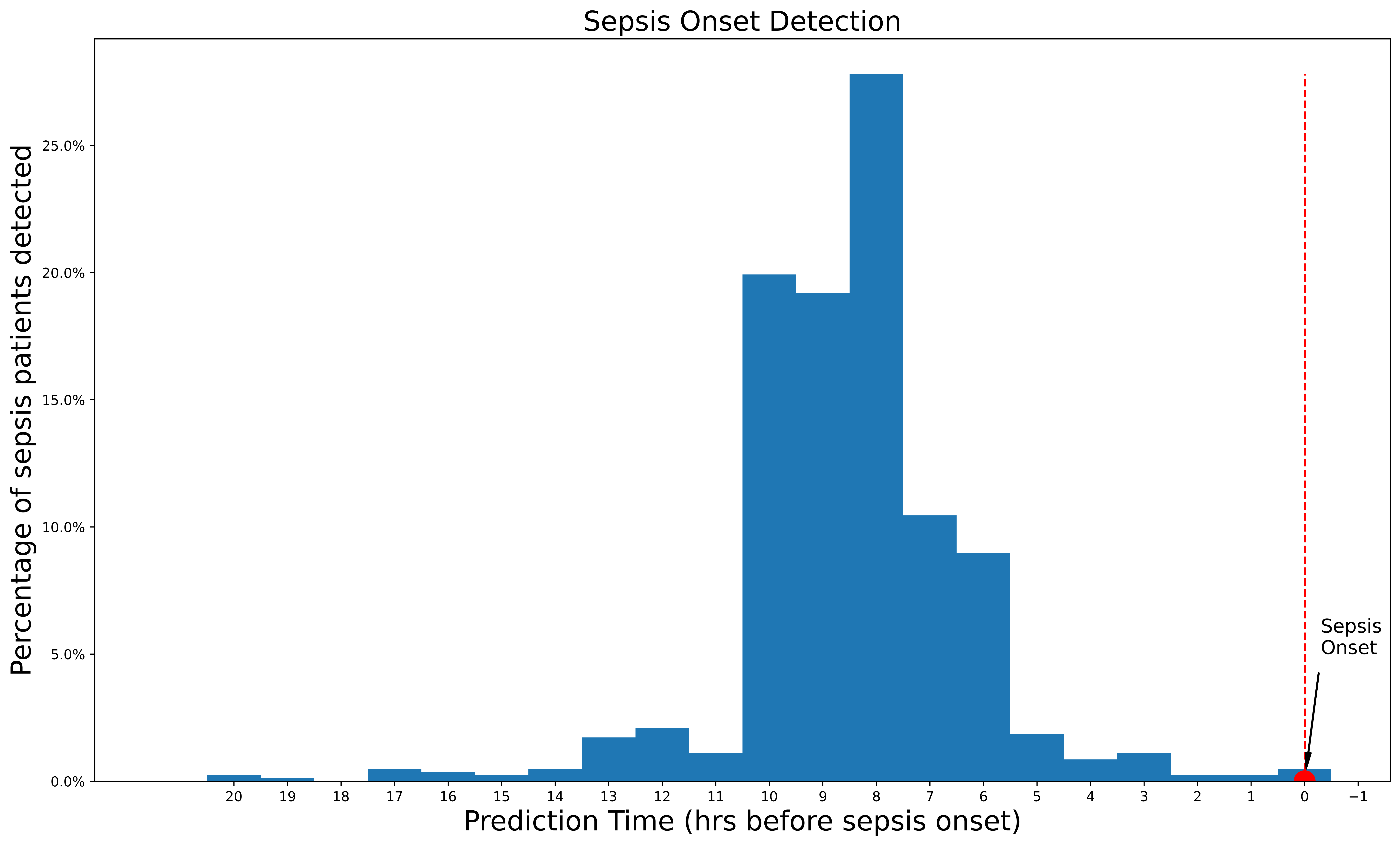}
  \caption{Model trained with Trust-MAPS}
  \label{fig:sub1}
\end{subfigure}%
\begin{subfigure}{.5\textwidth}
  \centering
  \includegraphics[width=\linewidth]{Figures/PredictionTimePercent_imp.png}
  \caption{Model trained without Trust-MAPS}
  \label{fig:sub2}
\end{subfigure}
\caption{We obtain the above plot by calculating how many hours in advance our ML model correctly predicts the onset of sepsis for every sepsis patient in the test set. Here, the time of sepsis prediction is the end of the \textit{sub-patient} time window at which our model raises the first sepsis alert for the patient. It is desirable to predict sepsis 6-12 hours before onset. Raising an alert too soon may be an outcome of a highly sensitive model prone to false positives, which can lead to alarm fatigue.}
\label{fig:time_to_pred}
\end{figure}

In Figure  \ref{fig:time_to_pred}, we plot a histogram depicting the distribution of the number of hours before sepsis onset that our algorithm raises its first alert for a patient. In Figure  \ref{fig:sub1}, we observe that most correct sepsis predictions were made $8-9$ hours before the patient developed sepsis, which is a {\it significant} lead time for clinical action. In Figure  \ref{fig:sub2}, we observe that the model trained without Trust-MAPS has a more widely spread distribution of sepsis detection times, which is not a desirable property for a clinical decision support model. A model that raises alerts for sepsis well before the prediction window of $6-12$ hours can be prone to false positives and can lead to alarm fatigue in the clinicians, whereas sepsis alerts raises too close to the time of onset may not provide enough time to take timely action.

\begin{table}[H]
\centering
\resizebox{\textwidth}{!}{ 
\begin{tabular}{|p{5.5cm}|c|c|c|c|c|}  
\hline
\textbf{Method} & \textbf{Utility} & \textbf{Accuracy} & \textbf{F-score} & \textbf{AUROC} & \textbf{AUPRC} \\ 
\hline
Trust-MAPS preprocessing (Normal Proj.) with  \cite{yang2020explainable} & \textbf{0.566 ±0.005} & \textbf{0.889 ±0.004} & \textbf{0.194 ±0.005} & \textbf{0.905 ±0.013} & \textbf{0.357 ±0.007} \\
\hline
Trust-MAPS preprocessing (Physical Proj.) with  \cite{yang2020explainable} & 0.565 ±0.004 & 0.882 ±0.065 & 0.187 ±0.004 & 0.903 ±0.003 & 0.355 ±0.011 \\
\hline
Baseline  \cite{yang2020explainable} (Without Trust-MAPS) & 0.522 ±0.004 & 0.845 ±0.078 & 0.151 ±0.005 & 0.894 ±0.003 & 0.193 ±0.018 \\
\hline
\end{tabular}
}
\caption{ Performance results with confidence intervals for the accuracy, AUC-ROC (Area under the Receiver Operating Characteristics Curve), AUC-PRC (Area under the Precision-Recall Curve), utility score ( \cite{Reyna2020}) and $f$-score on sepsis prediction combining Trust-MAPS as a preprocessing tool with the prediction algorithm proposed by Yang et al.  \cite{yang2020explainable}. }
\label{tab:benchmark_sota}
\end{table}

The \texttt{PhysioNet} Challenge  \cite{Reyna2020} used a \textit{utility score} metric to evaluate sepsis prediction models, which accounts for the timing between predicted and actual sepsis onset. This score penalizes predictions that are significantly early or late. To validate Trust-MAPS as a flexible preprocessing module compatible with any machine learning pipeline, we compared its performance against the sepsis prediction model developed by Yang et al. \cite{yang2020explainable}, which used XGBoost with Bayesian Optimization and achieved the highest average utility scores of 0.522 on the \texttt{PhysioNet} dataset. First, we replicate their pipeline to establish a baseline, then apply it to data processed with Trust-MAPS. We achieved an average utility score of 0.566—an 8\% improvement over the baseline of  \cite{yang2020explainable}. The results are documented in Table  \ref{tab:benchmark_sota}. We can clearly see a boost in performance over the baseline. 

The Supplemental contains details of our experiments with other classification algorithms (Section  \ref{app:res},  \ref{app:nosmote}), different sub-patient durations (Section \ref{app:different_time_window_sec}), and alternate imputation strategies (Section  \ref{app:alt_imputation}). 


\subsubsection{Feature Importance and Model Interpretability} \label{Interpretability}
A trained XGBoost model automatically calculates features of the maximum importance for the predictive modeling problem. Feature importance gives us insights into the clinical variables that have the maximum discriminatory power for sepsis prediction. It is calculated as the number of times a variable is selected for splitting a node, weighted by the squared improvement to the model as a result of each split, and averaged over all trees  \cite{elith2008working}. 
We calculate the average feature importance for each variable averaged over the $6$ hours sub-patient time window for our analysis. In Figure  \ref{fig:feature_imp_1}, we extend our analysis of feature importance to models trained on each of the 25 clusters identified by the k-means clustering step of our algorithm design, and plot the top 10 most important features for the prediction problem for each cluster. We find that the trust-scores were some of the most critical predictors in our ML model and imperative for effective decision-making. Trust-scores confer a natural interpretability to model predictions. This is highlighted in Table  \ref{tab:clinical}, where a trained clinician adjudicated the basis for sepsis classification for each cluster based on the variable specific trust-scores that were most important for classification in each cluster (Figure  \ref{fig:feature_imp_1}).

\begin{figure}[h]
    \centering
    \includegraphics[width = \textwidth]{Figures/Feature_imp_14dec.pdf}
    \caption{This figure shows the cluster-wise features of importance for XGBoost for each cluster identified by k-means clustering. The trust-scores (labeled "distance" variables), generated by calculating the distance of a sub-patient from the set of normal (or healthy) constraints, are marked in orange. We can see that these trust-scores play an important role in teaching the model to predict sepsis.}
    \label{fig:feature_imp_1}
\end{figure}

\begin{table}[!h]
    \centering
    \footnotesize 
    \begin{tabular}{|c|p{14 cm}|} \hline
    \textbf{Cluster} & \textbf{Clinical Interpretation}  \\ \hline 
       0  & Classification driven by subtle multiorgan derangement, with liver, SIRS and kidney predominance  \\ \hline
1	& Classification driven by subtle multiorgan derangement with predominant metabolic disturbance (glucose derangement) \\ \hline
2 &	Classification driven by subtle multiorgan derangement, with respiratory, hemodynamic (vascular tone) and inflammatory predominance \\ \hline
3&	Classification driven by subtle multiorgan derangements, with respiratory and liver predominance\\ \hline
4&	Classification driven by subtle multiorgan derangements, with hemodynamic predominance from changes in vascular tone \\ \hline
5&	Classification driven predominantly by inflammatory derangements (SIRS) \\ \hline
6&	Classification driven by subtle multiorgan derangements, with respiratory and inflammatory predominance\\ \hline
7	& Classification driven by subtle multiorgan derangements, with inflammatory and liver predominance \\ \hline
8&	Classification driven by subtle multiorgan derangements, with coagulopathy predominance\\ \hline
9&	Classification driven by subtle multiorgan derangements, with inflammatory and acidosis predominance\\ \hline
10&	Classification driven by subtle multiorgan derangements, with respiratory predominance due to oxygen measurement\\ \hline
11&	Classification driven by subtle multiorgan derangements, with inflammatory and kidney predominance\\ \hline
12&	Classification driven by subtle multiorgan derangements, with SIRS, kidney, and metabolic disturbance predominance \\ \hline
13	& Classification driven by subtle multiorgan derangements, with respiratory predominance due to oxygen requirements\\ \hline
14&	Classification driven predominantly by inflammatory derangements (temperature)\\ \hline
15	& Classification driven by subtle multiorgan derangements, with SIRS predominance\\ \hline
16	& Classification driven by subtle multiorgan derangements, with SIRS and respiratory predominance \\ \hline
17&	Classification driven by subtle multiorgan derangements, with respiratory failure, SIRS and liver predominance\\ \hline
18&	Classification driven by subtle multiorgan derangements, with repiratory predominance and concommitant acidosis\\ \hline
19	& Classification driven by subtle multiorgan derangements, with coagulopathy, metabolic (glucose) and inflammatory predominance\\ \hline
20&	Classification driven by subtle multiorgan derangements, with inflammatory dysequilibrium coagulopathy predominance\\ \hline
21	& Classification driven by subtle multiorgan derangements, with respiratory, kidney, metabolic (acidosis), and hemodynamic (vascular tome) predominance\\ \hline
22 &	Classification driven by subtle multiorgan derangements, with liver, acidosis, and respiratory predominance \\ \hline
23	& Classification driven by subtle multiorgan derangements, with inflammatory (temperature), respiratory failure, and liver failure\\ \hline
24&	Classification driven by subtle multiorgan derangements, with inflammatory, respiratory, acidosis, and coagulopathy predominance\\ \hline
    \end{tabular}
    \caption{The clinical interpretation relating to organ dysfunction profiles of the clusters using their most predictive features (i.e., specifically the trust-scores for various lab values and vitals).
    }
    \label{tab:clinical}
\end{table}

\section{Discussion}\label{sec:Discussions}

We introduce Trust-MAPS, a method for preprocessing EMR data that integrates clinical domain knowledge into data-driven pipelines using principles of constrained optimization. We demonstrate the framework’s utility by training a supervised machine learning classifier on temporal patient data, encompassing clinical laboratory values and vital signs. In healthcare applications, where predictive algorithms alert providers to potential medical conditions, the positive predictive value is crucial. Too many false positives could lead to system disengagement by healthcare staff, a challenge with current ML algorithms in practice  \cite{clifford2020future}. Therefore, we consider precision an essential metric for classifier performance. The model trained with Trust-MAPS achieved an AUC of 0.91 (95\% CI: 0.89–0.92) and a precision of 0.95 (95\% CI: 0.94–0.95) for predicting sepsis six hours before onset. This represents a substantial improvement over the baseline model trained on data without Trust-MAPS, which achieved an AUC of 0.79 (95\% CI: 0.77–0.81) and a precision of 0.50 (95\% CI: 0.49–0.51). Additionally, a model trained after applying only the first step of the Trust-MAPS process (the physical projection) achieved an AUC of 0.88 (95\% CI: 0.86–0.90) and a precision of 0.91 (95\% CI: 0.90–0.92) in the same prediction setting. 

Our results underscore two key points. First, the "physical projection" step alone improves classification performance, reinforcing the bias reducing capability of the error-handling step. Second, the trust-scores calculated by the "normal projection" serve as clinically interpretable features that enrich the feature space with predictive power, resulting in superior classification performance.

Clinical decision support models that accurately assess sepsis risk can be instrumental in reducing poor outcomes  \cite{islam2023machine}. As a result, many sepsis prediction algorithms have been developed across diverse data types and sources. For example, the authors in  \cite{goh2021artificial} developed an ML algorithm achieving an AUC of 0.90 and a precision of 0.82 for sepsis prediction 6 hours before onset. However, their results were based on a different dataset and incorporated clinical notes. To better assess Trust-MAPS, we benchmark our results against approaches that report outcomes on the \texttt{PhysioNet} Challenge dataset. The winning team  \cite{physionet1} of the challenge reports an average training set utility score of 0.430 and a test set utility score of 0.360. However, this comparison is limited, as the test utility scores were calculated on a hidden data set that we did not have access to. Therefore, we validate Trust-MAPS by testing it against the method proposed in  \cite{yang2020explainable}, which uses XGBoost with Bayesian Optimization to train a sepsis prediction model, achieving an average utility of 0.522. Higher predictive performance has been reported in works such as  \cite{gupta2024improving, strickler2023exploring, rosnati2021mgp}, but these approaches rely on deep learning. Their training pipelines, data sampling strategies,
and hyperparameter optimization methods are often underspecified and unavailable for reproducibility. In contrast, Yang et.  al.  \cite{yang2020explainable} provide a well-documented training pipeline and data sampling approach, enabling precise benchmarking. Using Trust-MAPS to preprocess data within this pipeline improves the final utility score to 0.566, underscoring that Trust-MAPS can serve as a modular preprocessing tool for robust machine learning.

The concept of using projections on high-dimensional mathematical domain constraints has important implications in practice. First, the projection onto physical constraints can help notify practitioners if specific feature measurements have recurring errors. A model with the capabilities to identify erroneous clinical data could have different purposes depending on the stage of model development or implementation, to serve as a generalized preprocessing pipeline, or to identify potential data drifts or manual entry errors. Projections onto clinical constraints can enrich most algorithms with the required clinical context to recognize \textit{potential} errors automatically. These outliers in EMR data can act as sources of systematic bias in ML models trained on this data, often disproportionately affecting certain patient populations or clinical scenarios and leading to unfair or inaccurate predictions. By enforcing clinically valid constraints through physical projections, Trust-MAPS actively mitigates these biases. Though the medico-legal implications of AI as diagnostic tools remains undefined  \cite{cestonaro2023defining}, our model's ability to compute projections on outlier data points could be treated as yet another quality control component in the healthcare infrastructure, particularly to guard against potential biases that may introduce generalization gaps during model development or implementation.

Second, trust-scores (or distance-to-normal measures) introduced in this work form a cohesive way to interpret key insights about patient physiology during the acute clinical trajectory. This approach can give clinicians or care providers the ability to quantify a patient's deviation from healthy physiology in a given interval of time, accounting for measured clinical values as well as temporal trends. By relying on fundamental physiological constraints rather than purely statistical normalization, our approach reduces the risk of learning spurious correlations that might reflect institutional practices or measurement biases rather than true clinical relationships. Trust-scores are an important meta-data that is predictive in the downstream ML  pipeline. In addition, we found that trust-scores have clinical relevance across the different clusters, as detailed in Table  \ref{tab:clinical}. The top seven vitals and laboratory data with the highest variance across all clusters are: TroponinI, PTT, Lactate, SaO$_2$, SOFA score, Bilibrubin-direct, and FiO$_2$. Our results are consistent with the work of  \cite{zhao2021early}, which reports PTT as a pivotal predictor of diagnosing sepsis early. The concept of trust-scores also aligns with contemporary clinical practice. The absolute value of a measurement is less important than how much it deviates from normal. This is demonstrated in Figure  \ref{fig:feature_imp_1} in which a trust-score is the most important feature used to differentiate sepsis from non-sepsis in 17 of the 25 clusters. For instance, upon encountering a patient with multi-organ failure such as one experiencing significant liver failure (cluster 0, Figure  \ref{fig:feature_imp_1}), the degree to which an organ is failing is usually in direct proportion to how deviant the variable representing that organ is from "normal". A total bilirubin of 24 mg/dL is farther than 1.2 mg/dL (the upper limit of normal for total bilirubin) than 14 mg/dL. Thus, this projection-based framework can enable similar levels of targeted and precise clinical management that may not be possible when considering models within the strong interpretation component afforded by such pipelines.

There are some limitations to this work. First, while we use the Euclidean distance metric to compute projections, alternative metrics may better suit certain constraints. Additionally, all constraints for clinical features were designed specifically for ICU patients in consultation with practicing clinicians; tailoring these constraints to specific patient sub-groups could potentially enhance performance and enable precision medicine. Further, all constraints are equally weighted in projection calculations, though assigning lower weights to frequently erroneous variables may improve error handling. Another limitation is that Trust-MAPS has been tested on a single dataset; extending this approach to additional datasets with potential domain shifts would help evaluate its generalizability. Finally, our pipeline can compute projections and run the machine learning model only after the full sub-patient data sequence (six hours with three-hour overlap) is available, introducing a three-hour delay for real-time predictions. Shortening the sub-patient sequence duration could reduce this lag but might limit insights derived from temporal patterns in the data.

Leveraging projections to enhance data processing pipelines remains a relatively unexplored area of research. This work represents an initial investigation into this concept and demonstrates how incorporating clinically valid constraints can support health equity by ensuring models make predictions based on physiologically plausible data rather than artifacts or biases in the collection process. In future work, we aim to investigate other distance metrics (e.g. KL divergence) for our projections pipelines. The constraints on clinical data can also be modified according to the illness being modeled, or the cohort being studied. Alternate formulations of the optimization objective could potentially lead to more faithful data correction results, for example, a stronger penalty on rate of change of measures variables. Further, when creating sub-patients, we set the sepsis label for the \textit{sub-patient} to `1' (yes) if the original patient ultimately developed sepsis in the time window considered for that sub-patient. In future work, it would be interesting to consider other design choices, for example, setting the \textit{sepsis label} to be a continuous variable between 0-1, to show gradual progression to sepsis.

\section{Conclusion}\label{sec13}

Predictive models that integrate data from various sources are highly sensitive to errors in data. Rapid identification of errors in clinical data is paramount to reducing alarm fatigue, improving robustness, and building unbiased models  \cite{vellido2018machine}.  

Trust-MAPS is a novel technique grounded in constrained optimization principles that we propose for enriching machine learning pipelines for clinical decision support. It can help with outlier detection and management in EMR data, and can engineer a new set of features (trust-scores) with high predictive power for training machine learning models. We demonstrate the utility of this tool within the context of sepsis prediction for critically ill patients in the Intensive Care Unit (ICU). The proposed Trust-MAPS pipeline addresses the timely need to develop a new, context-aware framework that can dynamically identify nonsensical data, model the domain by translating clinical expertise into mathematical constraints, use projection-based methods to automatically handle discrepancies in data to highlight trustworthy data as well as potential errors. We develop Trust-MAPS as a tool that can enrich any machine learning algorithm with clinical context, mitigating bias and increasing trustworthiness.

We finally remark that our projection approaches are general. The proposed concept of using projections to detect and correct erroneous data can be helpful in other clinical settings or non-clinical domains. We hope the positive results of this study inspire other researchers to use projections for data corrections and trust-scores to enhance their prediction pipelines.

\paragraph{Data Availability} \label{sec:data_avail}
The data analyzed in this study is from the \texttt{PhysioNet:} Computing in Cardiology Challenge 2019 and is publicly available at \url{https://physionet.org/content/challenge-2019/1.0.0/}. The code used for our preprocessing and imputation pipeline, projection pipeline, and machine learning pipeline is documented and available at \url{https://github.com/Kamaleswaran-Lab/ClinicalProjections_SepsisOnset}.

\clearpage
\bibliographystyle{IEEEtran}
\bibliography{export.bib}
\newpage

\pagenumbering{arabic}

\newpage
{\Large \centering \textbf{Supplementary File}}
\section{Methods}\label{sec:methods}

\begin{figure}[h]
    \includegraphics[width=\textwidth, trim={0 3cm 0 3cm},clip]{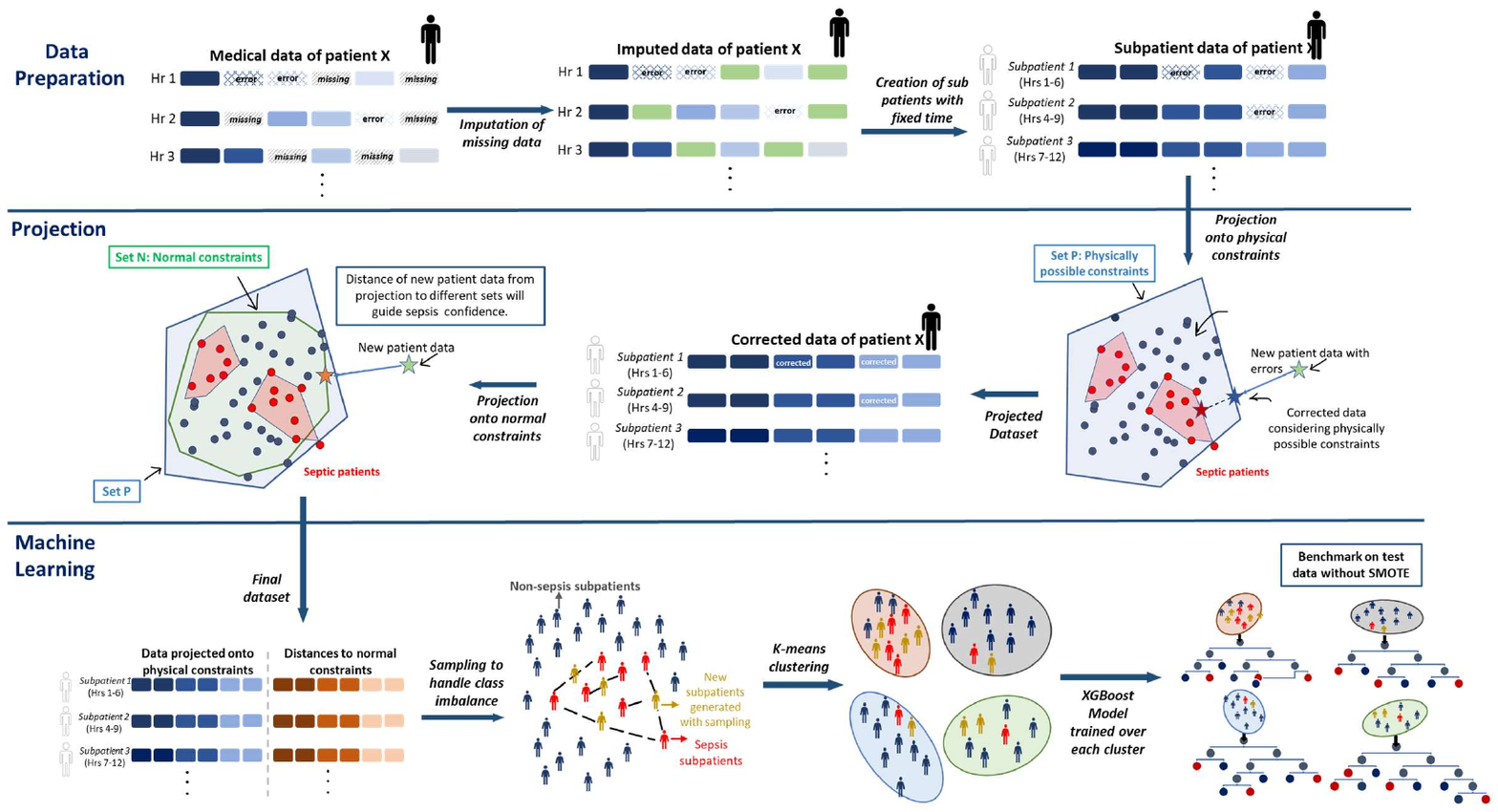}
    \caption{\footnotesize A schema of Trust-MAPS following the data preparation, projection, and machine learning steps, for sepsis prediction. (a) Our algorithm uses clinical laboratory values and vitals of ICU patients recorded over their ICU stay, denoted by $f_i$ in the image. (b) We impute missing values in patient records according to the strategy described in Section \ref{sec:prep}. (c) We create fixed dimension \textit{sub-patient} records of six-hour time windows with three-hour overlaps for all patient encounters. (d) We first correct erroneous data by the physical projection onto the set $P$, defined as the set of biologically possible constraints that any patient's data must adhere to. (e). Next, we obtain ``error-corrected'' \textit{sub-patient} records. (f) We then calculate the distance of these projected data points to the Set $N$, defined as the set of normal constraints consisting of the clinical data normal/healthy ranges. (g) This gives us trust-scores for our ML prediction algorithm, which we appended to each \textit{sub-patient} record. (h) We perform Synthetic Minority Oversampling Technique (SMOTE) \cite{chawla2002smote} on the training set to upsample the minority class (sepsis) to overcome class imbalance. (i) We then perform K-means clustering on the training data, as described in Section \ref{sec:ML}. (j) We train the XGBoost (eXtreme Gradient Boosting) decision trees \cite{chen2016xgboost} over each cluster as described in Section \ref{sec:ML}. (k) We benchmark our approach by comparing the performance of a similar machine learning algorithm on data with the application of Trust-MAPS, and data without the application of Trust-MAPS.}
    \label{fig:pipeline}
\end{figure}

In this section, we outline our algorithmic pipeline in detail, which is composed of three steps: (i) preprocessing and imputation (Section \ref{sec:prep}), (ii) projections  onto domain constraints using Trust-MAPS (Section \ref{sec:projections}), and (iii) machine learning via clustering (Section \ref{sec:ML}). Although we include one clustering-based approach for predicting sepsis in Section \ref{sec:ML}, the technique of data correction using projections can be used in conjunction with other learning approaches. We refer the reader to Figure \ref{fig:pipeline} for a visual representation of our algorithmic pipeline, detailing different data dimensions and augmentation. 

\subsection{Data Preparation and Imputation Pipeline} 
\label{sec:prep}

The first step in our algorithmic pipeline is the preprocessing and imputation pipeline, composed of three sub-steps: (i) imputation (Section \ref{sec:impute}), (ii) normalization (Section \ref{sec:norm}), and (iii) standardization of time intervals (Section \ref{sec: stand}). We report the physical and normal upper and lower bounds for the different clinical variables we consider in this work in Table \ref{tab:ranges} in Section \ref{app:bound}.

\subsubsection{Imputing Missing Values} \label{sec:impute}
 
Missing clinical variable values in every patient encounter are imputed before proceeding with the proposed projections pipelines. Missing value imputation is used quite frequently in many validated algorithms for sepsis prediction \cite{nemati2018interpretable, moor2021early, yang2020explainable}. We use linear interpolation to fill in missing values between two recorded measurements for any clinical variable except FiO$_2$, for which we use fill-forward (step-wise) interpolation. If missing values cannot be interpolated because initial measurement values are missing, we fill them in with the mean of the normal (expected) range of values for healthy patients (refer Table \ref{tab:ranges} in Section \ref{app:bound}), except for FiO$_2$, which we set to a default value of $21$.    

\subsubsection{Normalization} \label{sec:norm}
We normalize the imputed data to ensure that variables with large magnitude values do not skew the subsequent projections. We utilize min-max normalization to scale all clinical variables down, using the upper and lower bounds of the normal range ($N$). This means that recorded measurements below the lower bound of the normal range would map to negative values and measurements greater than the upper bound of the normal range would map to values greater than $1$. For example, since the normal range of heart rate is 60-90: a heart rate of $54$ in the original would map to $\frac{54-60}{90 - 60} = -0.2$ after normalization, a heart rate of $120$ would map to $\frac{120-60}{90-60} = 2.0$.

Although normalization scales the data down, some medical variables can still have high values even after normalization because their physically possible ranges ($P$) can vary by many orders of magnitude from the normal range. To mitigate this issue, we log-transform ($\text{Log-transform}(x) = \log_{10} (x + 1)$) variables for which the width of the physical range is at least two orders of magnitude greater than the normal range lower bound, first, before normalizing them\footnote{We add $1$ to the original value of the variable $x$ to avoid applying the log transform on zeros.}. For example, the physical range of Creatinine is 0-20 (the width of that range is $20$), whereas its normal range is 0.5-1.3, so we log-transform it. The log-transformed variables are then normalized using their log-transformed upper and lower normal bounds, thus mapping variables that have a high deviation from the normal range onto the logarithmic scale comparable to the other normalized variables. For example, a Creatinine value of 14 maps to $\frac{\log_{10}(1 + 14) - \log_{10}(1 + 0.5)}{\log_{10}(1 + 1.3) - \log_{10}(1 + 0.5)} = 5.39$. We refer the reader to {Table \ref{tab:ranges} in Supplemental Materials section \ref{app:bound}}, where we list the variables that we log-transformed.

\subsubsection{Standardizing the Length of Time Intervals} \label{sec: stand}

\begin{figure}
    \centering
    \includegraphics[width=0.8\linewidth]{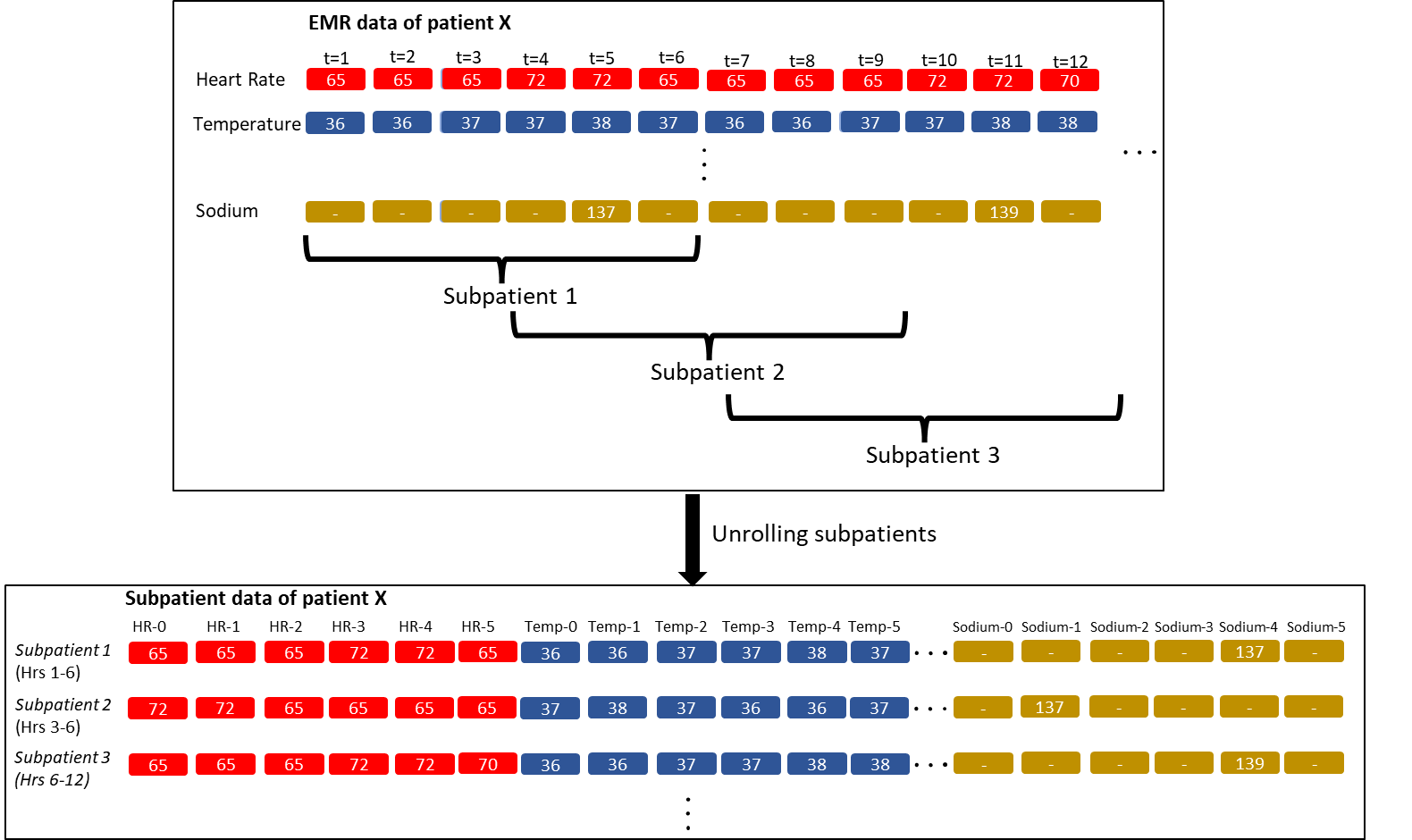}
    \caption{Visualization of the \textit{sub-patient} creation process. For this work, we use $6$ hour time intervals with $3$ hours of overlap.}
    \label{fig:subpatientCreation}
\end{figure}

To deal with varying length-of-stays in the ICU and standardize the data dimensions for applying Trust-MAPS and subsequent learning, we divide each patient’s data into overlapping segments of fixed time intervals, called \textit{sub-patients}. A \textit{sub-patient} can be considered a snapshot-in-time of the patient's entire length of stay in the ICU. For a fixed time interval $n$, and number of clinical variables $f$, we take snapshots of size $n \times f$ of patient data, unroll each clinical feature, and create one dimensional sub-patients of length $n*f$. Each \textit{sub-patient} is then treated as a separate datapoint input to our data processing and machine learning pipeline. A visualization of the \textit{sub-patient} creation process is shown in Figure \ref{fig:subpatientCreation}. 

We set the sepsis label for the \textit{sub-patient} to ``1'' (septic) if the original patient developed sepsis anywhere in the time window considered for that sub-patient\footnote{One can consider other design choices, for example, setting the \textit{sepsis label} to be a continuous variable between 0-1, to show gradual progression to sepsis. We leave this design choice as future work.}.

For predicting clinical diseases like sepsis, \textit{sub-patients} with larger time intervals provide a greater longitudinal context of the progression of a patient's clinical condition. However, \textit{sub-patients} with shorter observation time-intervals are useful as they make predictions using only the most recent patient data. Therefore, the prediction algorithm in real-time is robust to spurious ICU events. We use overlapping windows to create \textit{sub-patients} to better capture forward and backward time context of trends in a patient's clinical data. In our analysis, we compare the performance of different time window lengths and overlap times and include the results in Supplemental Materials section \ref{app:different_time_window_sec}. For the purpose of results and discussions in this work, we choose a fixed time interval of 6 hours, with 3-hour overlaps. 

\subsection{A Novel Projection Pipeline} \label{sec:projections} 

The key idea is to use projections to correct errors at the sub-patient level, incorporate ``trust'' in the patient data, and quantify how sick a patient is. This information is used downstream in an ML pipeline to predict acute conditions like sepsis. Precisely, we will model the ``feasibility'' of clinical data by using mathematical constraints obtained by translating clinical knowledge (e.g., heart rate cannot be below 30 for a living patient), and projections to the resultant region will automatically correct some errors in given clinical data for any patient. The distance of the input data from the projected data will be used to form a ``trust-score'' in the data; in particular, we will also quantify how sick a patient is by projecting to a set $N$ that models control physiology. The sets we project clinical data on will lie in a high-dimensional space and are non-continuous and non-convex (which itself is a computationally challenging task). Our projections pipeline is composed of two steps that we describe in what follows: (i) projection onto physically possible constraints $P$ (Section \ref{sec:phys proj}); (ii) projection onto normal constraints $N$ (Section \ref{sec:norm proj}). The constraints were placed on the vitals and lab values by a panel of three board-certified physicians (co-authors A.L.H., P.Y., J.W.). Two physicians (P.Y. and J.W.) independently reviewed the full dataset by investigating the normal distribution of each data variable. Through this process, they characterized each clinical variable into a feasible and infeasible range, identifying potential erroneous data elements based on established physiological norms. Discrepancies in data categorization were resolved through consensus with a third investigator (A.L.H).

\subsubsection{Projections onto the Physically Possible Constraints} \label{sec:phys proj}
The first clinical decision set $P$ that we consider is defined by constraints on vitals and laboratory values to which any patient's data must adhere. In other words, if the data does not satisfy these constraints, then the data must have errors. We use $\dindex{vital}$ to denote the {\it pre-processed} (imputed, normalized, and log-transformed if applicable) value of a clinical variable 
for sub-patient $i$ and time step $t$, and $\pindex{vital}$ for its projected value that we wish to compute. 

We specify physical lower and upper bounds on each vital for each time period in the fixed six-hour time interval for each \textit{sub-patient}, as well as constrain the rates of change of the clinical data per hour in Table \ref{tab:ranges} in Supplemental Materials section \ref{app:bound}. These are translated into constraints on the closest-point optimization problem as follows:

\begin{itemize}[leftmargin= 50 pt]
    \item[\it Constraint 1:] All vitals must satisfy their physical range lower and upper bounds (e.g., temperature lies in [25, 45] when the unit is Celsius). 
    \item[\it Constraint 2:] All applicable vitals must adhere to the maximum physical change per hour (e.g., temperature can change at most by two units within an hour)
\end{itemize}

These constraints are mathematically written as follows \footnote{In practice, we normalize and log-transform the bounds and rates used in the constraints to match our data preprocessing. For readability, we present the constraints here using values prior to normalized and log-transforms.} for each \textit{sub-patient} $i$, of a fixed window length $n$ ($n=6$ for experiments reported in the paper). 
\begin{align*}
    25 &\leq \pindex{Temp} \leq 45  \quad \forall t \in \{1,\dots, n\}, \\
    -2 &\leq \pindex{Temp} - x[\text{Temp},i,t-1] \leq 2 \quad  \forall t \in \{2,\dots, n\}.
\end{align*}

Our method is general enough to consider more complex clinical constraints involving more variables. For example, we can model if-then-else constraints as follows:

\begin{itemize}[leftmargin= 50 pt]
    \item[\it Constraint 3:] If HCO$_3 \leq 10$, then BaseExcess $\leq$ 0,
\end{itemize}

Mathematically, this can be written using binary variables $z_i^{(t)}$ to model the if-then-else logic as follows:
\begin{equation*}
    \begin{aligned}
       \pindex{HCO_3} & \leq 10z_i^{(t)} + 45(1 - z_i^{(t)}), \\
       \pindex{HCO_3} &\geq 10(1 -  z_i^{(t)}),\\
       \pindex{BaseExcess} &\leq 20(1 -  z_i^{(t)})\\
      z_i^{(t)} &\in \{0,1\},
      \\
    &\hspace{-50 pt} \text{for all } t \in \{1,\dots, n\} \text{ and each sub-patient } i. 
    \end{aligned}
\end{equation*}
To verify the mathematical modeling of \textit{Constraint 3}, consider when $z_i^{(t)} = 1$. Then the above constraints reduce to
\begin{equation*}
    \begin{aligned}
        \pindex{HCO_3}  &\leq 10, \\  
        \pindex{HCO_3} &\geq 0, \\
        \pindex{BaseExcess} &\leq 0,
    \end{aligned}
    \end{equation*}
which implies that $\pindex{HCO_3} \leq 10$ and $\pindex{BaseExcess} \leq 0$. Similarly, when $z_i^{(t)} = 0$ we have
\begin{equation*}
    \begin{aligned}
        \pindex{HCO_3} &\leq 45, \\ \pindex{HCO_3}&\geq 10, \\ \pindex{BaseExcess} &\leq 20,
    \end{aligned}
\end{equation*}
which implies that we are in the other case when $ \pindex{HCO_3} \geq 10$. We have no restrictions on $\pindex{BaseExcess}$ and $\pindex{HCO_3}$. Indeed, the remaining constraints $\pindex{HCO_3} \leq 45$ and $\pindex{BaseExcess} \leq 20$ are redundant since 45 and 20 are the physically possible upper bounds for HCO$_3$ and BaseExcess, respectively. Thus we obtain the desired outcome based on how the value of $z_i^{(t)}$ is decided during the projection computation.

The remaining constraints that we consider to define the decision set $P$ are stated next, while their mathematical formulation is included in the Supplemental Materials section \ref{app:proj}. 

\begin{itemize}[leftmargin= 50 pt]
    \item[\it Constraint 4:] Relationship between MAP, DBP, and SBP must be satisfied approximately at any time $t$: \\
\begin{align*}
\mathrm{MAP} &\geq 0.95 \left(\frac{2}{3} \mathrm{DBP} + \frac{1}{3} \mathrm{SBP}\right) \quad\\
&\text{and}\\
\quad \mathrm{MAP} &\leq 1.05 \left(\frac{2}{3} \mathrm{DBP} + \frac{1}{3} \mathrm{SBP} \right).
\end{align*}

    \item[\it Constraint 5:]
    At any time $t$, if Lactate $\geq 6$, then BaseExcess $ \leq 0$.
    
    \item[\it Constraint 6:] At any time $t$, if BaseExcess  $\leq$ 0, then either Lactate $\geq 6$ or {HCO$_3$} $\leq 10$.
    
    \item[\it Constraint 7:] At any time $t$, if $\mathrm{pH} \leq 7$, then either $\mathrm{PaCO_2} \leq 35$ or HCO$_3 \leq 10$.
    
    \item[\it Constraint 8:] Relationship between $\mathrm{BilirubinDirect}$ and $\mathrm{BilirubinTotal}$ at any time $t$:\\ $\mathrm{BilirubinDirect} \leq \mathrm{BilirubinTotal}$.
    
\item[\it Constraint 9:] Relationship between $\mathrm{HCT}$ and $\mathrm{Hgb}$ at any time $t$:
$\mathrm{HCT} \geq 1.5\mathrm{Hgb}$.
\end{itemize}

Letting $V$ denote the set of clinical variables (listed in Table \ref{tab:ranges}) under consideration, the computation of a projection onto any set $Q$ is then mathematically defined as
\begin{equation}\label{physical proj}
    \min_{x \in Q} \sum_{v \in V} \sum_{i \in I} \sum_t (\dindex{v} - \pindex{v})^2.
\end{equation}

We let $x_Q^*$ denote the optimal solution of \eqref{physical proj} in what follows. We first set $Q$ to be equal to $P$, which is the (non-convex) decision set arising from the intersection of all the \textit{constraints (1-9), and the type of vital values (continuous or binary)}. We include the complete mathematical formulation in Supplemental Materials section \ref{app:proj}. A solution to the mathematical problem \eqref{physical proj} for $Q$ equal to $P$, which we call the \textit{physical projection}, finds ``corrected'' data with the least total squared change compared to the given data and recovers linear imputation in many cases. 

The output $x_P^*$ of the above optimization problem handles outliers in the data that lie outside the physical range or do not satisfy the physiological feasibility constraints, by bringing them to the closest point that satisfies these constraints. We use  \textit{physical projection} and \emph{corrected dataset} interchangeably to denote $x_P^*$.

\subsubsection{Projection onto Homeostasis Constraints and Trust-Scores} \label{sec:norm proj}

The second clinical decision set $N$ we consider is the \textit{control physiology set} defined by constraints composed of normal ranges on vitals and laboratory values that any healthy patient's data must adhere to. Let $U_{v}^N$ and $L_{v}^N $ denote the normal upper and lower bounds for a vital $v \in V$, respectively. 

The constraints defining $N$ ensure that all vitals satisfy their normal lower and upper bounds. Thus, for sub-patient $i$ and time step $t$ in a fixed window length $n$ ($n=6$ for results presented in the paper),

\begin{itemize}[leftmargin= 80 pt]
    \item[\it Constraints for N:] $L_{v}^N \leq \pindex{v} \leq U_{v}^N$ for all $t \in \{1,\dots n\}$, \textit{sub-patient} $i$, and vitals $v \in V$.
\end{itemize}
For example, in the case when the vital $v$ is temperature, we have the following constraints:
\begin{align*}
    36 &\leq \pindex{Temp} \leq 38, \\
    &\hspace{15 pt}\text{for all } t \in \{1,\dots, n\} \text { and each sub-patient }i.
\end{align*}

We now let the decision set $Q$ be the set $N$ and solve the optimization problem \eqref{physical proj} using data points $x_P^*$ generated after projecting onto $P$. We call the optimal solution $x_N^*$ the \textit{normal projection}. The distance of vital $v$, for sub-patient $i$, from the decision set $N$ is what we refer to as the ``trust-score'' of vital $v$ for that particular sub-patient. The higher $\text{NormDist}(v, i)$ are, the sicker the patient in the time interval of the \text{sub-patient}.

\begin{equation} \label{NormDistance}
    \text{NormDist}(v,i) \coloneqq \sum_{t = 1}^6(\dindex{{v}} - x_N^*[v, i,t])^2
\end{equation}

Next, we normalize $\text{NormDist}(v,i)$ across subpatients using a standard min-max normalization. We call these newly normalized normal distances ``trust-scores'' as they incorporate the clinical domain measure of how sick a patient is, allow transparency and a natural explainability in machine learning model predictions, and (as shown in Section \ref{sec: ml results}) enrich the feature space to learn better classifiers and obtain better precision, thereby increasing trust in these artificial intelligence systems. We then append these (normalizes) normal distances for all vitals $v \in V$ to our \emph{corrected} dataset. The final dimension of each \textit{sub-patient} data is $ \mid I \mid \times (\mid V \mid + \mid V\mid  + 3 + 2 )$. The additional $3$ comes from including age, gender, and patient ID (to be used for patient identification and descriptive statistics), and the next additional 2 comes from the including SIRS scores, and SOFA scores, which we computed for each sub-patient.

\subsection{Machine Learning Pipeline} \label{sec:ML}
We now describe the machine learning (ML) pipeline for predicting sepsis and elaborate on important design choices that improved predictive performance for this task. 

\paragraph{Train-Test Split:} The learning pipeline is designed in a way that overcomes class imbalance and ensures that the results that we report are fair representations of the true performance of our model. To ensure this, before application of any resampling technique, we split our dataset into a training and testing set in a 72:25 ratio, at the patient level. We take care to split the sepsis and non-sepsis patients separately, each class being split in a 75:25 ratio. This is done so that the results we see on the test set are not skewed because of a severely imbalanced split. The process is repeated for every cross-validation run of the learning pipeline.

\paragraph{SMOTE:}  The first challenge in training a machine learning model for our use-case was overcoming class imbalance. To overcome this issue, we use an artificial dataset up-sampling technique called Synthetic Minority Oversampling Technique (SMOTE) \cite{chawla2002smote}. The method uses random convex combinations of observed data points belonging to the minority class to synthesize the new data points, thereby populating the minority class with synthesized data that is relatively close in the feature space to the observed data \cite{bookimbalanced}. SMOTE has been used in earlier studies that develop machine learning classifiers for other clinical conditions such as oral cancer detection \cite{carnielli2018combining}, cell identification/classification \cite{xia2020machine}, and sepsis detection \cite{goh2021artificial} where the prevalence of the positive cases are low. 
Recent works have computationally shown that combining random undersampling of the majority class with minority class upsampling using SMOTE performs better than only using SMOTE on the minority class \cite{chawla2002smote}. In our case, we combine random undersampling of the majority class with upsampling for the minority class using SMOTE on 
to obtain a $25:75$ ratio of sepsis to non-sepsis data points on the dataset we use for training the model.

We refer the reader to {Figure \ref{fig:UMAP_sepsis_nonsepsis} in the Supplemental Materials section,} which shows a UMAP (Uniform Manifold Approximation and Projection \cite{mcinnes2018umap}) for a two-dimensional visualization of the data for sepsis and non-sepsis patients after the application of SMOTE.

\paragraph{Clustering:}  Given the highly heterogeneous nature of patient data, we follow the Cluster-then-Predict approach \cite{ClusterPredict} to maximize performance. 

As the first step in our learning process, we resort to unsupervised $k$-means clustering methods to help us find trends in the data and understand which patient vitals contribute the most to sepsis diagnosis. In addition, clustering allows us to tailor our supervised learning methods by training a separate machine learning model for each cluster following the Cluster-then-Predict approach \cite{ClusterPredict} to maximize performance. Letting $n$ denote the number of points in our training dataset $\mathcal{X}^{\mathcal{D}}$, the $k$-means clustering gives us a mapping $h: \mathcal{X}^{\mathcal{D}} \mapsto \{1, \dots, k\}^n$ from our training dataset to a grouping of \textit{sub-patients} into clusters. This mapping $h$ groups \textit{sub-patients} with a similar feature space together. To choose the number of clusters $k$, we looked at two metrics; (i) the ``elbow'' of the mean-squared error curve; (ii) the maximum within cluster concentration as a function of the number of clusters.\footnote{That is, for a given number of clusters $k$, we computed the percentage of septic patients within each cluster and took the maximum overall $k$ clusters.} 

\paragraph{Supervised Learning:} For each cluster $c \in \{1, \dots, k\}$, we train a machine learning classifier $M_c$ on data points from the SMOTE-upsampled training data $\mathcal{X}^{\mathcal{D}}$ mapped to $c$ using the mapping $h$, i.e., the k-means clustering. At the time of testing, the same mapping $h$ is used to map \textit{sub-patients} to $c \in \{1, \dots, k\}$, and the machine learning classifier $M_c$ is used for prediction. We use XGBoost (eXtreme Gradient Boosting) decision trees \cite{chen2016xgboost} to train the final classifiers for each cluster. To maximize precision, we threshold predicted probabilities of the classifier using the value that maximized the $f$-score (Equation \ref{fscore}) for the training data. These thresholds are given in Figure \ref{fig:thresholds} in Supplemental Materials section \ref{app:comp}.
\begin{equation}
    f\text{-score} = \frac{2 \times Precision \times Recall}{Precision + Recall}.
    \label{fscore}
\end{equation}

We choose $f$-score as our metric as it finds a good balance between precision and recall for the prediction task and improves our positive predictive value.

We benchmark the performance of our algorithm by running $50$ iterations with a different train-test split. We ensure that the test set at each iteration is split at the patient level and does not have SMOTE applied to it. The same cluster-based machine-learning pipeline is run on data that is not processed using Trust-MAPS, and the performance of the two algorithms are compared in Table \ref{tab:results1}.

\newpage
\section{Formulation of Physical Projection Model} \label{app:proj}
Recall that we use $V$ to denote the set of all vitals in Table \ref{tab:ranges} that have a physical range (i.e. all vitals except age and gender) and $I$ to denote the index-set of sub-patients under consideration in the \texttt{imputed dataset} ($\mathcal{I}$). We let $H$ denote the set of vitals in Table \ref{tab:ranges} that have a maximum hourly change. Further, we use $\dindex{v}$ to denote the actual value of vital $v \in V$ in $\mathcal{I}$ for sub-patient $i \in I$ and time interval $t$, and $\pindex{v}$ for its projected variable that we wish to compute. Recall that We use $V$ to denote the set of all vitals in Table \ref{tab:ranges} that have a physical range (i.e. all vitals except Age and gender) and $I$ to denote the index-set of sub-patients. Further, we let $H$ denote the set of vitals in Table \ref{tab:ranges} that have a maximum hourly change. Further, let $U_\mathrm{v}^P$, $L_\mathrm{v}^P$, and $R_\mathrm{v}^P$ denote the physical upper bound, physical lower bound, and the maximum hourly change (when applicable) for vital $v$ respectively. We can now write the physical projection problem through the following mixed-integer quadratic program (MIQP):

\begin{align*}
\min_{x} &~ \sum_{v \in V} \sum_{i \in I} \sum_{t=1}^6 (\dindex{v} - \pindex{v})^2 \\
\text{s.t.} &~ \intertext{\texttt{Constraint 1:  All vitals must satisfy their physical lower and upper bounds.}}
&L_v^P \leq \pindex{v} \leq U_v^P,  ~~\forall \; \text{$v\in V$, $t \in \{1,\dots 6\}$ and sub-patients $i \in I$.} 
&\intertext{\texttt{Constraint 2:  All applicable vitals must adhere to the maximum physical change per hour.}}
&-R_v^P \leq \pindex{v} - x[\text{v},i,t-1]  \leq R_v^P,  \forall \; \text{$v\in H$, $t \in \{2,\dots 6\}$ and sub-patients $i \in I$.} 
&\intertext{\texttt{Constraint 3:  If HCO$_3 \leq 10$, then BaseExcess $\leq$ 0.}}
& \pindex{HCO_3} \leq 10\gindex{z}{HCO_3} + 45(1 - \gindex{z}{HCO_3}),\\
& \pindex{HCO_3} \geq 10(1 -  \gindex{z}{HCO_3}),\\
& \pindex{BaseExcess} \leq 20(1 -  \gindex{z}{HCO_3}),\\
& \gindex{z}{HCO_3} \in \{0,1\}, ~~\forall \; \text{$t \in \{1,\dots 6\}$ and sub-patients $i \in I$.}
&\intertext{\texttt{Constraint 4:  Relationship between MAP, DBP and SBP:}}
& \pindex{MAP} \geq 0.95 \left(\frac{2}{3} \pindex{DBP} + \frac{1}{3} \pindex{SBP}\right), \\
& ~~\forall \; \text{$t \in \{1,\dots 6\}$ and sub-patients $i \in I$.}\\
&\pindex{MAP} \leq 1.05 \left(\frac{2}{3} \pindex{DBP} + \frac{1}{3} \pindex{SBP}\right), \\
& ~~ \forall \; \text{$t \in \{1,\dots 6\}$ and sub-patients $i \in I$.}
&\intertext{\texttt{Constraint 5:  If Lactate $ \geq 6$, then BaseExcess $\leq$ 0.}}
& \pindex{Lactate} \leq 30\gindex{z}{Lactate} + 6(1 - \gindex{z}{Lactate}), \\
& \pindex{Lactate} \geq 6\gindex{z}{Lactate}), \\
& \pindex{BaseExcess} \leq 20(1 -  \gindex{z}{Lactate}),\\
& \gindex{z}{Lactate} \in \{0,1\}, \forall \; \text{$t \in \{1,\dots 6\}$ and sub-patients $i \in I$.}
&\intertext{\texttt{Constraint 6:  If BaseExcess $\leq$ 0, then either {Lactate} $\geq 6$ or {HCO$_3$} $\leq 10$.}}
& \pindex{BaseExcess} \leq 20(1 -  \gindex{z}{BaseExcess}),\\
& \pindex{BaseExcess} \geq -40 \gindex{z}{BaseExcess}),\\
& \pindex{HCO_3} \leq 10\gindex{y}{HCO_3} + 45(1 - \gindex{y}{HCO_3}),\\
& \pindex{HCO_3} \geq 10(1 -  \gindex{y}{HCO_3}), \\
& \pindex{Lactate} \leq 30\gindex{s}{Lactate} + 6(1 - \gindex{s}{Lactate}), \\
& \pindex{Lactate} \geq 6\gindex{s}{Lactate}), \\
& \gindex{z}{BaseExcess} \leq \gindex{y}{HCO_3} +  \gindex{s}{Lactate}), \\
& \gindex{z}{BaseExcess}, \gindex{y}{HCO_3}, \gindex{s}{Lactate}) \in \{0,1\}, \\
& ~~\forall \; \text{$t \in \{1,\dots 6\}$ and sub-patients $i \in I$.}
\intertext{\texttt{Constraint 7: If $\mathrm{pH} \leq 7$, then either $\mathrm{PaCO_2} \leq 35$ or {HCO$_3$} $\leq 10$.}}
& \pindex{pH} \leq  7\gindex{z}{pH} +  7.7(1 -  \gindex{z}{pH}),\\
& \pindex{pH} \geq 7(1-\gindex{z}{pH}) + 6.5 \gindex{z}{pH}), \\
& \pindex{HCO_3} \leq 10\gindex{s}{HCO_3} + 45(1 - \gindex{s}{HCO_3}), \\
& \pindex{HCO_3} \geq 10(1 -  \gindex{s}{HCO_3}),\\
& \pindex{PaCO2} \leq 35\gindex{y}{PaCO2} + 120(1 - \gindex{y}{PaCO2}), \\
& \pindex{PaCO2} \geq 35(1-\gindex{y}{PaCO2}) + 16 \gindex{y}{PaCO2}, \\
& \gindex{z}{pH} \leq \gindex{s}{HCO_3} + \gindex{y}{PaCO2}),\\
& \gindex{z}{pH}, \gindex{s}{HCO_3}, \gindex{y}{PaCO2}) \in \{0,1\}, \\
& ~~\forall \; \text{$t \in \{1,\dots 6\}$ and sub-patients $i \in I$.} 
&\intertext{\texttt{Constraint 8:  Relationship between $\mathrm{BilirubinDirect}$ and $\mathrm{BilirubinTotal}$:}}
& \pindex{BilirubinDirect} \leq  \pindex{BilirubinTotal},\\
& ~~\forall \; \text{$t \in \{1,\dots 6\}$ and sub-patients $i \in I$.}
&\intertext{\texttt{Constraint 9:  Relationship between $\mathrm{HCT}$ and $\mathrm{Hgb}$:}}
& \pindex{HCT} \leq  1.5 \pindex{Hgb}, ~~\forall \; \text{$t \in \{1,\dots 6\}$ and sub-patients $i \in I$.}
\end{align*}  

\newpage 
\section{Data Description: Sepsis Labeling} \label{app:sepsis}
The sepsis onset times for patients as defined by the creators of the  \texttt{physioNet} Computing in Cardiology 2019 \cite{Reyna2020} Challenge is included in this section. The following time stamps were defined for each patient's data:

\begin{enumerate}
    \item \textbf{t\textsubscript{suspicion}} : The time of clinical suspicion of infection 
    \begin{itemize}
        \item  Identified as the earlier timestamp of IV antibiotics and blood cultures within a specified duration. If antibiotics were given first, then the cultures must have been obtained within 24 hours. If cultures were obtained first, then antibiotic must have been subsequently ordered within 72 hours. Antibiotics must have been administered for at least 72 consecutive hours to be considered.
    \end{itemize}
    
    \item \textbf{t\textsubscript{SOFA}} : The time of occurrence of end organ damage
    \begin{itemize}
        \item Identified by a two-point drop in SOFA score within a 24-hour period.
    \end{itemize}
    
    \item \textbf{t\textsubscript{sepsis}} : The onset time of sepsis 
    \begin{itemize}
        \item Identified as the earlier of \textbf{t\textsubscript{suspicion}} and \textbf{t\textsubscript{SOFA}} as long as \textbf{t\textsubscript{SOFA}} occurs no more than 24 hours before or 12 hours after \textbf{t\textsubscript{suspicion}}. Specifically, if 
        \begin{equation}
            \textbf{t\textsubscript{suspicion}} - 24 \leq \textbf{t\textsubscript{SOFA}} \leq \textbf{t\textsubscript{suspicion}}+12  \Rightarrow \textbf{t\textsubscript{sepsis}}=\min(\textbf{t\textsubscript{suspicion}},\textbf{t\textsubscript{SOFA}})
        \end{equation}
        
        \item If the above condition is not met, the patient is not marked as a sepsis patient. 
    \end{itemize}
\end{enumerate}

\newpage
\section{Clinical Vitals Data} \label{app:bound}
First, we report the physical upper and lower bounds for different blood levels in Table \ref{tab:ranges}, and include bound and rate constraints wherever possible. 
\begin{table}[H]
    \centering
    \begin{tabular}{|p{3 cm}||p{3 cm}|p{3 cm}|p{3 cm}|} \hline
  \textbf{Patient Vital / Parameter}   & \textbf{Physiologically Plausible Range} & \textbf{Normal / Healthy Range} & \textbf{Max change within 1 hour} \\ \hline \hline 
  Heart rate & 30-200 & 60-90 & N/A \\ \hline
   O$_2$Sat & 50-100 & 95-100 & N/A \\
   \hline
    Temperature ($^{\circ}$C) rate & 25-45 & 36-38 & 2 \\ \hline
    SBP & 50-200 & 90-130 & N/A \\
    \hline
     MAP & 20-140 & 65-75 & N/A \\ \hline
     DBP & 20-150 & 60-80 & N/A \\ \hline
     Respiration & 8-70 & 10-24 & 40 \\ \hline
     ETCO$_2$ & 10-80 & 35-45 & 30 \\ \hline
     BaseExcess & $-40$-20 & $-2$-2 & 10 \\ \hline
     HCO$_3$ & 0-45 & 22-27 & N/A \\ \hline
     FiO$_2$ & 21-100 & $\leq 20$ & N/A \\ \hline
      pH & 6.5-7.7 & 7.35-7.45 & N/A \\ \hline
       PaCO$_2$ & 16-120 & 35-45 & N/A \\ \hline
    SaO$_2$ & 50-100 & 95-100 & N/A \\ \hline
    Chloride & 50-150 & 96-106 & N/A \\ \hline
    Creatinine* & 0-20 & 0.5-1.3 & N/A \\ \hline
    Bilirubin (direct)* & 0-50 & 0-0.4 & N/A \\ \hline
    Bilirubin (total)* & 0.5-80 & 0.2-1.2 & N/A \\ \hline
    Glucose* & 10-3000 & 60-200 & 300 \\ \hline
    Lactate* & 0-30 & 0.5-1 & N/A \\ \hline
    Magnesium & 0-10 & 1.5-2.5 & N/A \\ \hline
    Phosphate & 0-12 & 2.5-4.5 & N/A \\ \hline
     Potassium & 1-12 & 3.5-4.5 & N/A \\ \hline
     TroponinI* & 0-80 & 0-0.3 & N/A \\ \hline
     HCT & 10-50 & 35-45 & N/A \\ \hline
     Hgb & 0-17 & 12-17 & N/A \\ \hline
     PTT & 0-200 & 60-70 & N/A \\ \hline
     WBC* & 0-200 & 4-11 & N/A \\ \hline
     Fibrinogen & 0-999 & 200-400 & N/A \\ \hline
     Platelets & 0-1500 & 150-450 & N/A \\ \hline
      Age & 18-110 & N/A & 1 \\ \hline
       Gender & N/A & Male\textbackslash Female & N/A \\ \hline
\end{tabular}
    \caption{All the patient vitals/parameters that exist in the \texttt{PhysioNet} data along with their normal and physical ranges. The range values in the table are used in the constraints of our projections model. Further, starred vitals are those whose bounds for their physical ranges differ by at least two orders of magnitude and are thus log-transformed. All vitals with the exception of the last two are used in our projection models. 
    }
    \label{tab:ranges}
\end{table}

Next, we report the vital characteristics of the study population in Table \ref{tab:char}. The data of 40,336 patients was available to us, out of which  2,932 patients (7.2\%) develop sepsis. Then, a total of 463,693 sub-patients were created by our data processing strategy, out of which 9,646 (2\%) correspond to sepsis patients. The mean age of sepsis and non-sepsis sub-patients was 62. About 56\% of the sub-patients were males, with 59\% of the sepsis sub-patients being male. Comparing the laboratory values and vitals of the sepsis and non-sepsis sub-patient population did show statistically significant differences, as can be observed from Table \ref{tab:char}.

\begin{table}[h]
\centering
\begin{tabular}{|c||c|c|c|c|}
\hline
&     & \textbf{Overall}       & \textbf{No Sepsis}           & \textbf{Sepsis}    \\
\hline \hline
 $n$                             &           & 463693        & 454047        & 9646                     \\ \hline
 Age, mean (SD)              &             & 62.1 (16.4)   & 62.1 (16.4)   & 62.1 (16.3)        \\ \hline
 Gender, n (\%)
     & M          & 259306 (55.9) & 253588 (55.9) & 5718 (59.3)         \\
\hline
 SOFA*, mean (SD)             &             & 0.9 (1.1)     & 0.9 (1.1)     & 1.2 (1.3) \\ \hline
 HR, mean (SD)                    &        & 84.4 (16.3)   & 84.3 (16.3)   & 90.4 (17.8)   \\ \hline
 Temp, mean (SD)                  &        & 36.9 (0.7)    & 36.9 (0.7)    & 37.1 (0.9)    \\ \hline
 MAP, mean (SD)                   &        & 82.6 (14.2)   & 82.6 (14.2)   & 80.6 (13.7)   \\ \hline
 DBP, mean (SD)                   &        & 65.0 (11.6)   & 65.0 (11.6)   & 63.6 (11.5)   \\ \hline
 BaseExcess, mean (SD)            &        & 0.0 (2.4)    & 0.0 (2.4)    & 0.1 (3.3)     \\ \hline
 HCO3, mean (SD)                  &        & 24.5 (3.0)    & 24.5 (3.0)    & 24.4 (3.7)    \\ \hline
 Lactate, mean (SD)               &        & 1.2 (1.1)     & 1.2 (1.1)     & 1.5 (1.5)     \\ \hline
 PaCO2, mean (SD)                 &        & 40.3 (6.4)    & 40.3 (6.4)    & 40.5 (8.5)    \\ \hline
 pH, mean (SD)                    &        & 7.4 (0.0)     & 7.4 (0.0)     & 7.4 (0.1)     \\ \hline
 Hct, mean (SD)                   &        & 32.0 (5.6)    & 32.0 (5.6)    & 31.0 (5.5)    \\ \hline
 Hgb, mean (SD)                   &        & 10.7 (2.0)    & 10.7 (2.0)    & 10.4 (2.0)    \\ \hline
 BilirubinDirect, mean (SD)      &        & 0.3 (1.0)     & 0.3 (1.0)     & 0.4 (1.6)     \\ \hline
 BilirubinTotal, mean (SD)       &        & 1.1 (2.2)     & 1.0 (2.1)     & 1.5 (3.4)     \\ \hline
 O2Sat, mean (SD)                 &        & 97.1 (2.5)    & 97.1 (2.5)    & 96.9 (2.8)    \\ \hline
 SBP, mean (SD)                   &        & 123.6 (20.6)  & 123.6 (20.6)  & 121.6 (21.4)  \\ \hline
 SaO2, mean (SD)                  &        & 96.0 (5.9)    & 96.0 (5.9)    & 95.4 (6.5)    \\ \hline
 Resp, mean (SD)                  &        & 18.7 (4.3)    & 18.7 (4.3)    & 20.2 (5.6)    \\ \hline
 EtCO2, mean (SD)                 &        & 39.4 (4.3)    & 39.4 (4.3)    & 38.6 (5.3)    \\ \hline
 FiO2, mean (SD)                  &        & 42.0 (298.4)  & 41.7 (301.3)  & 55.0 (70.4)   \\ \hline
 Creatinine, mean (SD)            &        & 1.4 (1.7)     & 1.4 (1.7)     & 1.6 (1.7)     \\ \hline
 Glucose, mean (SD)               &        & 130.3 (41.6)  & 130.2 (41.5)  & 135.3 (45.3)  \\ \hline
 Magnesium, mean (SD)             &        & 2.0 (0.3)     & 2.0 (0.3)     & 2.1 (0.4)     \\ \hline
 Phosphate, mean (SD)             &        & 3.4 (1.1)     & 3.4 (1.1)     & 3.5 (1.3)     \\ \hline
 Potassium, mean (SD)             &        & 4.1 (0.5)     & 4.1 (0.5)     & 4.1 (0.6)     \\ \hline
 TroponinI, mean (SD)             &        & 1.1 (8.7)     & 1.1 (8.7)     & 1.3 (9.9)     \\ \hline
 PTT, mean (SD)                   &        & 49.5 (21.1)   & 49.5 (21.0)   & 46.9 (22.8)   \\ \hline
 WBC, mean (SD)                   &        & 11.0 (6.6)    & 10.9 (6.5)    & 12.6 (8.2)    \\ \hline
 Fibrinogen, mean (SD)            &        & 303.2 (63.9)  & 303.0 (63.0)  & 312.5 (96.7)  \\ \hline
 Platelets, mean (SD)             &        & 208.0 (101.8) & 208.0 (101.4) & 208.9 (117.5) \\ \hline
 Chloride, mean (SD)              &        & 103.4 (4.6)   & 103.4 (4.5)   & 104.1 (5.4)   \\
\hline
\end{tabular}
    \caption{Sub-patient characteristics grouped by the sepsis label on the \texttt{imputed dataset}. The class imbalance can be observed from the first row of the data-set where only around 2\% of sub-patients have sepsis. *SOFA score was partially computed with available data.}
    \label{tab:char}
\end{table}

The data of 40,336 patients was available to us, out of which  2,932 patients (7.2\%) develop sepsis. Then, a total of 463,693 sub-patients were created by our data processing strategy, out of which 9,646 (2\%) correspond to sepsis patients. The mean age of sepsis and non-sepsis sub-patients was 62. About 56\% of the sub-patients were males, with 59\% of the sepsis sub-patients being male. Comparing the laboratory values and vitals of the sepsis and non-sepsis sub-patient population did show statistically significant differences, as can be observed from Table \ref{tab:char}.


\clearpage
\section{More Computational Results} \label{app:comp}

\subsection{Physical Projection Plots}

\begin{figure}[H]
    \centering
    \includegraphics[width = \textwidth]{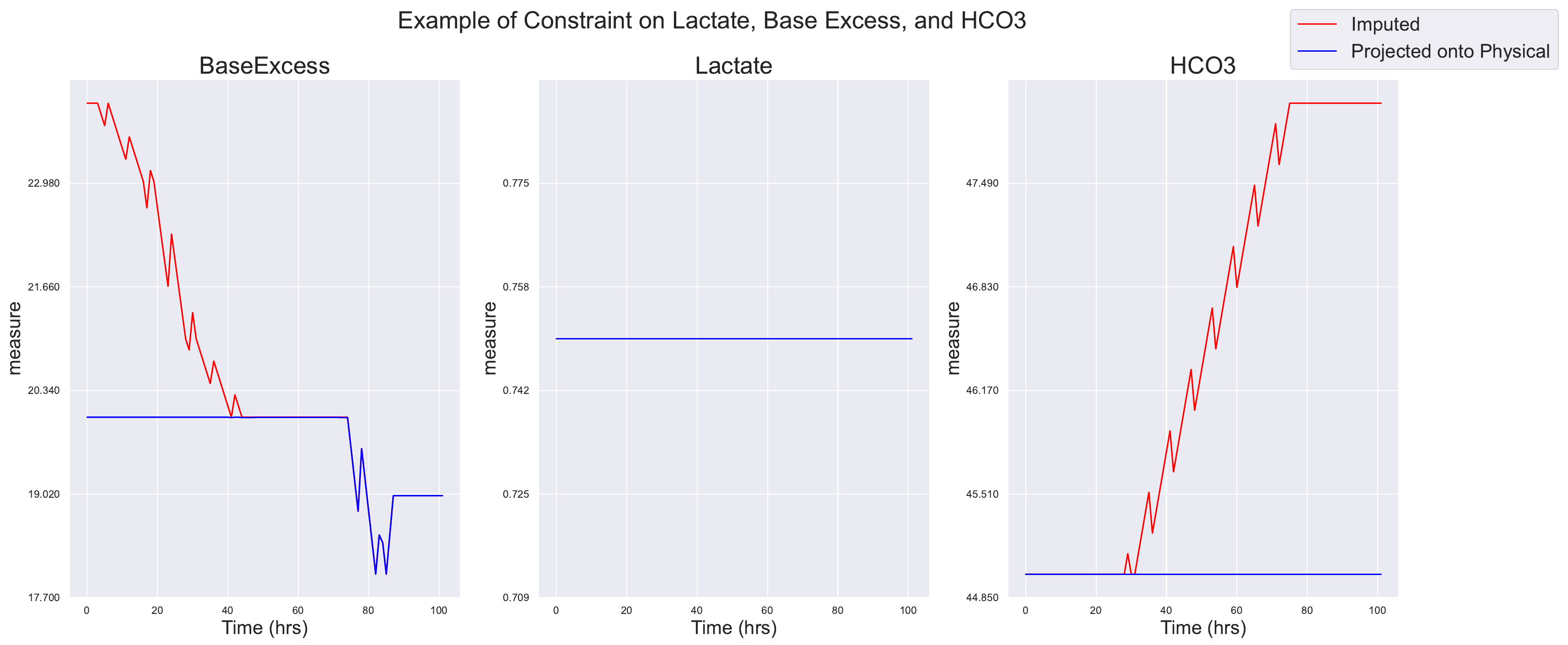}
    \caption{Another illustrative example of a the behavior of the physical projection demonstrated through the constraint \texttt{\#6} on the logical relationship between BaseExcess, Lactate and HCO$_3$.}
    \label{fig:proj behav 4}
\end{figure}

First, we give examples showing the behavior of how the physical projection corrects our EMR data in Figures \ref{fig:proj behav 3} and \ref{fig:proj behav 4}:

\begin{figure}[H]
    \centering
    \includegraphics[scale = 0.16]{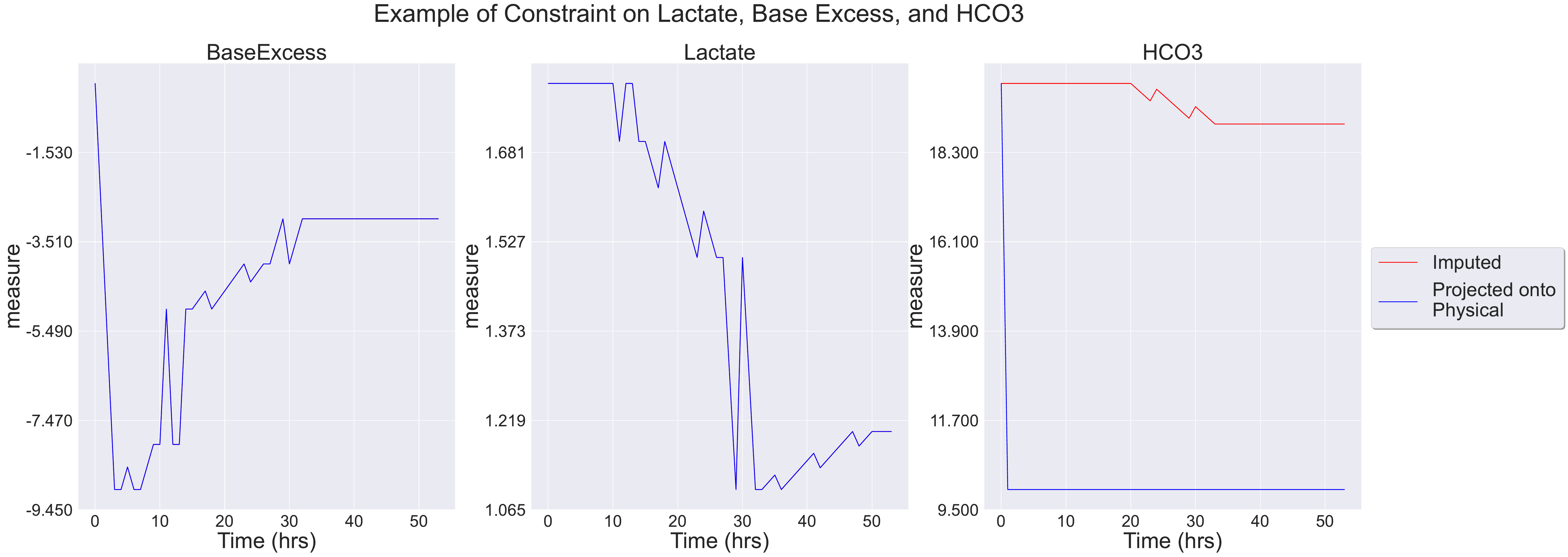}
    \caption{An illustrative example of a the behavior of the physical projection demonstrated through the constraint \texttt{\#6} on the logical relationship between BaseExcess, Lactate and HCO$_3$.}
    \label{fig:proj behav 3}
\end{figure}

\begin{figure}[h]
    \centering
    \includegraphics[width=\textwidth]{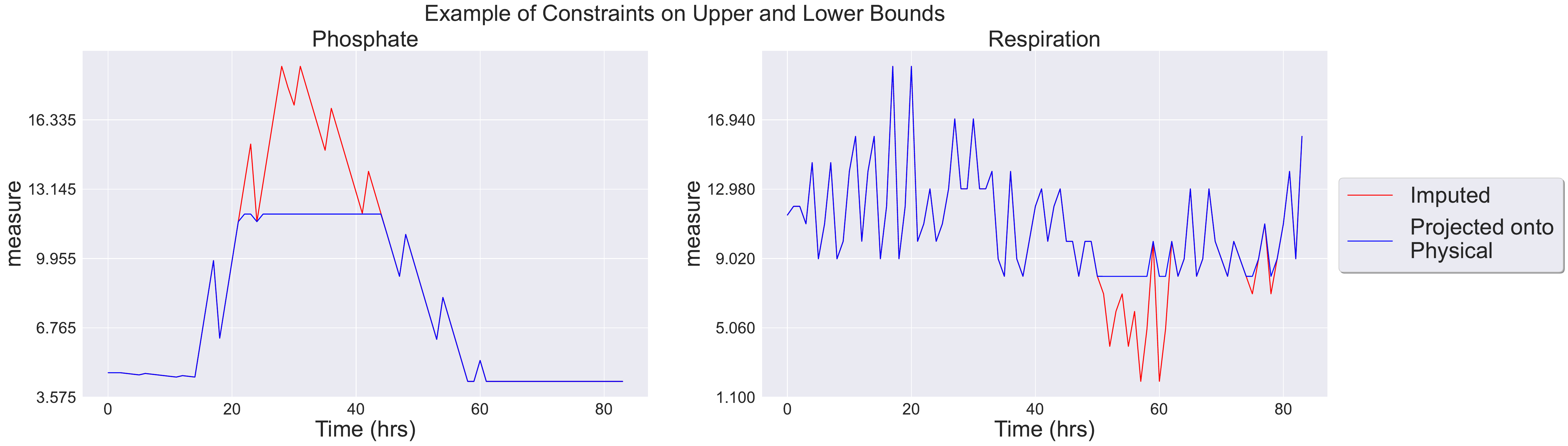}
    \caption{Example of Upper and Lower Bound Constraints}
    \label{fig:ulbounds}
\end{figure}

\begin{figure}[h]
    \centering
    \includegraphics[width=0.5\textwidth]{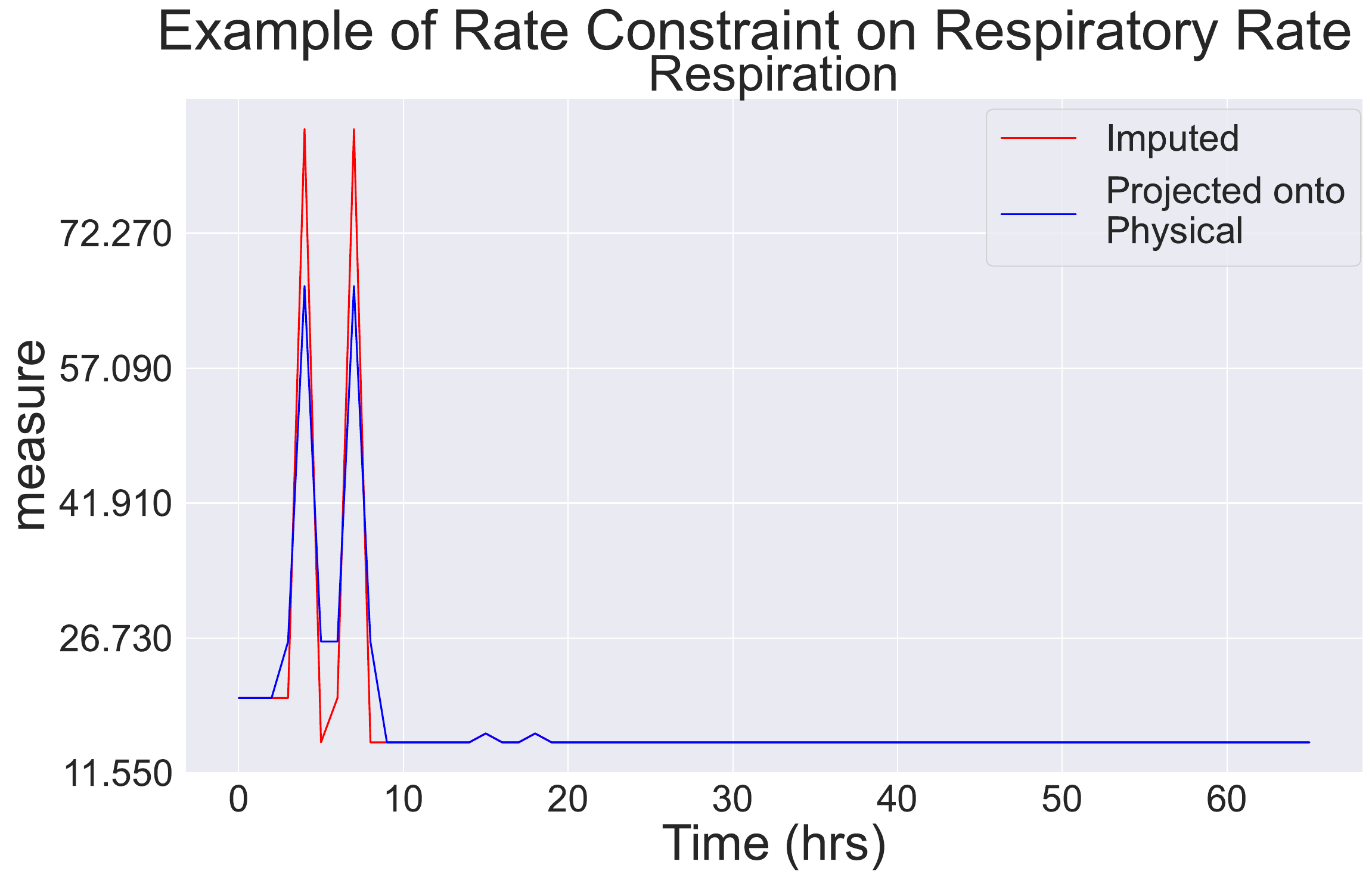}
    \caption{Example of rate constraints on respiratory rate}
        \label{fig:resp}
\end{figure}

\subsection{Clustering Plots}
\begin{figure}[htbp]
    \centering
    \includegraphics[scale = 0.5]{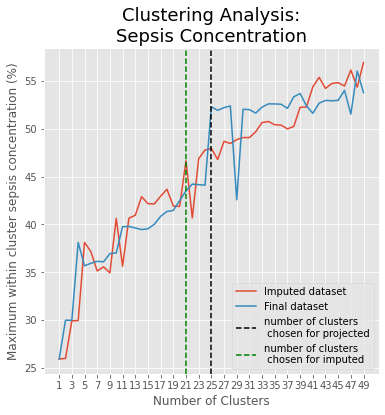}
    \includegraphics[scale = 0.5]{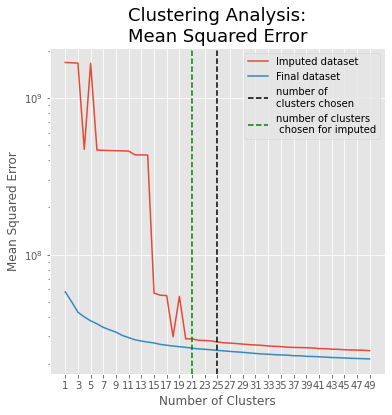}
    \vspace{-5 pt}
    \caption{\textbf{Left:} Plot showing the maximum within cluster sepsis concentration as a function of the number of clusters specified when running the $k$-means algorithm for the \texttt{imputed dataset without Trust-MAPS} applied ($\mathcal{X}^{\mathcal{I}}_S$) v/s \texttt{final dataset with Trust-MAPS} applied ($\mathcal{X}^{\mathcal{F}}_S$). \textbf{Right:} Plot showing the mean squared error (MSE) of the $k$-means algorithm for the imputed v/s projected data-sets, where we see that the projected data has a lower MSE even though its a bigger data-set. The final number of clusters chosen for imputed and projected data-sets are given by the dashed lines and was determined by looking at the elbow of the MSE curve and the maximum within cluster sepsis concentration.}
    \label{fig:clustering} 
    \vspace{-5 pt}
\end{figure}

We next present clustering plots comparing the results with and without projections. First, we compare the performance of the $k$-means algorithm on the projected and imputed datasets in Figure \ref{fig:clustering}, and find that the mean squared error on the projected data (on subpatients) is much less than the mean squared error on the imputed data set.

\begin{figure}[htbp]
    \centering
    \includegraphics[scale = 0.3]{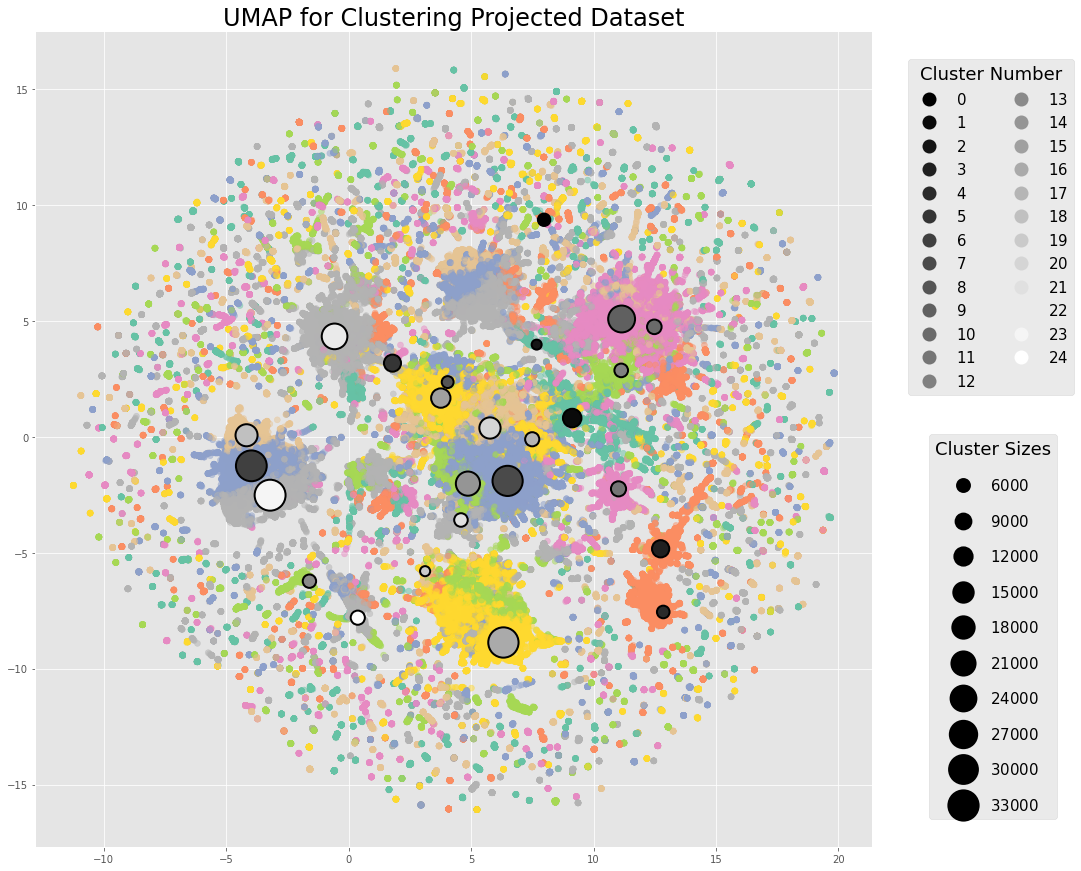}
    \caption{UMAP Plot showing the clustering results on \texttt{projected dataset with Trust-MAPS} after SMOTE resampling ($\mathcal{X}^{\mathcal{F}}_S$).}
    \label{fig:UMAP1}
\end{figure}

\begin{figure}[htbp]
    \centering
    \includegraphics[scale = 0.35]{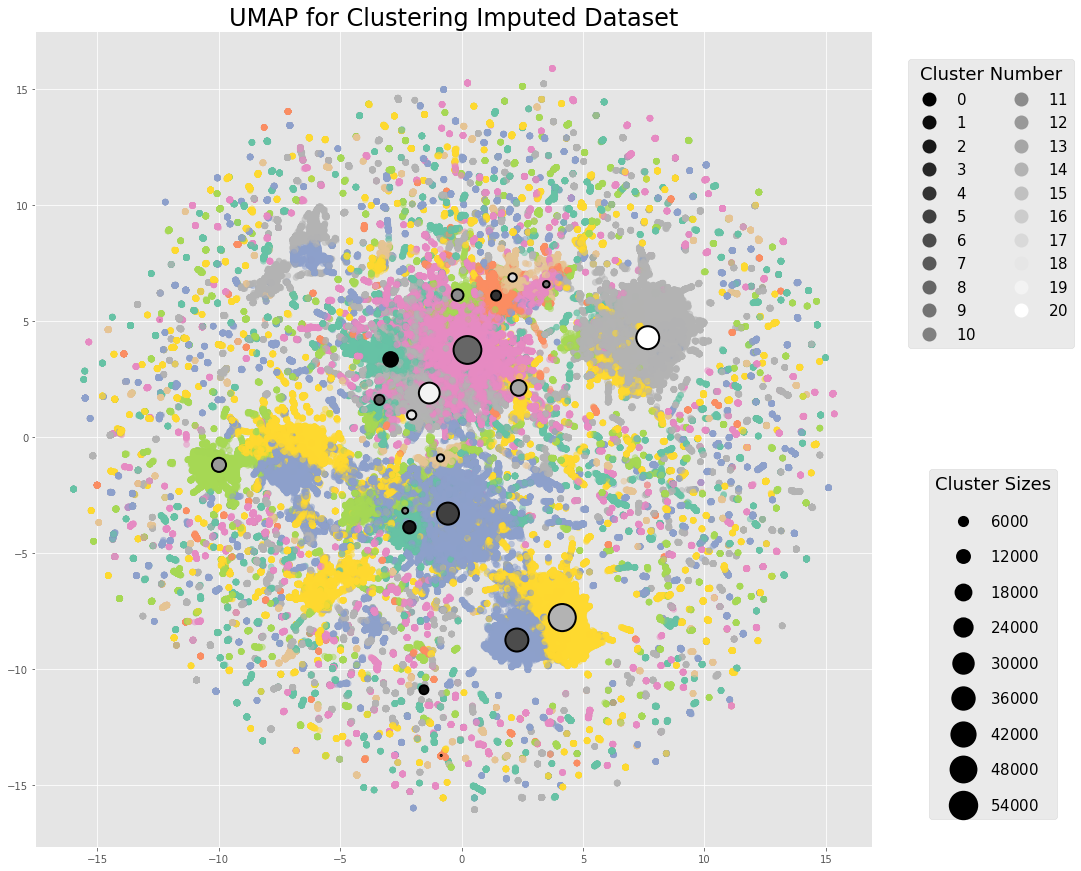}
    \caption{UMAP Plot showing the clustering results on \texttt{imputed dataset without Trust-MAPS} after SMOTE resampling ($\mathcal{X}^{\mathcal{I}}_S$).}
    \label{fig:UMAP2}
\end{figure}

\begin{figure}[htbp]
    \centering
    \includegraphics[scale = 0.25]{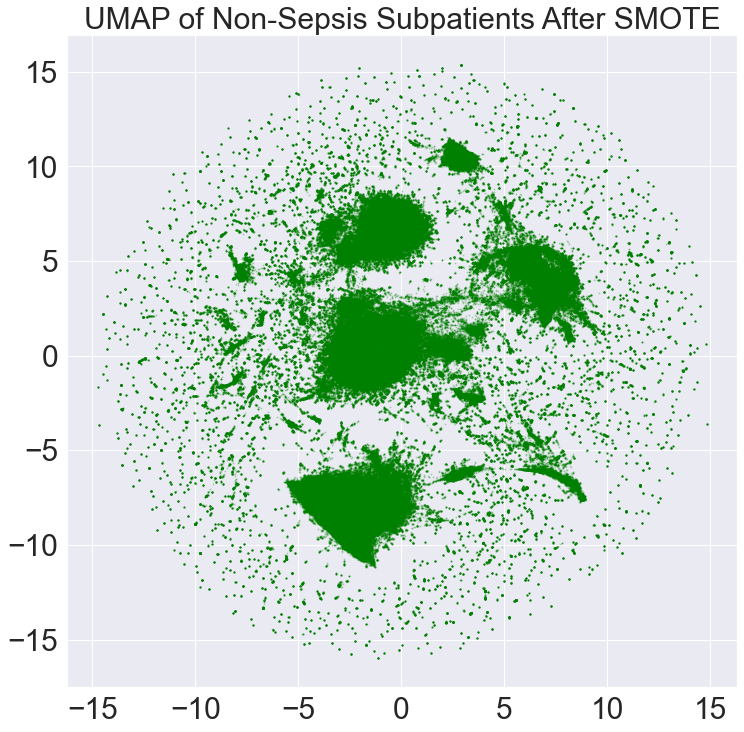}
    \includegraphics[scale = 0.25]{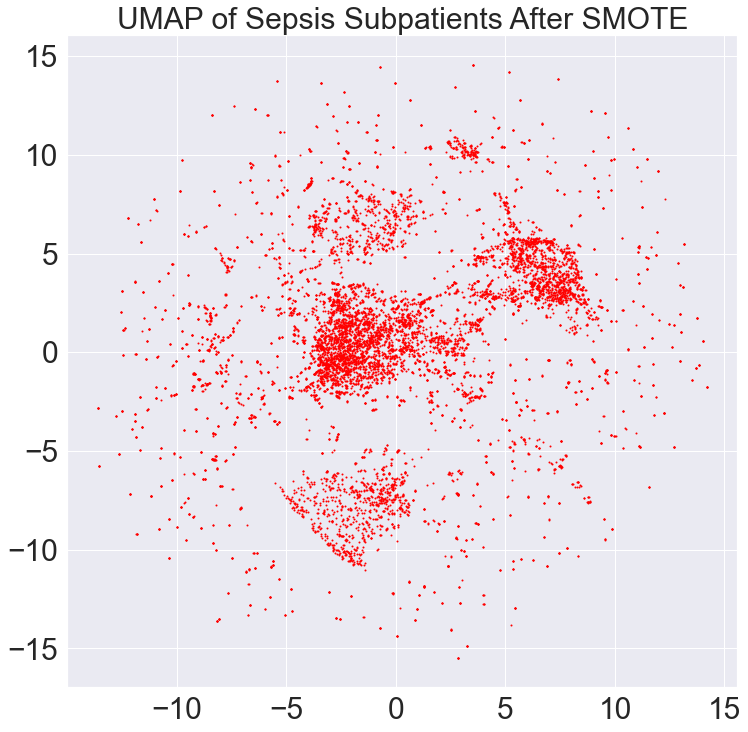}
    \caption{\textbf{Left.} UMAP plot showing the distribution of non-sepsis sub-patients in the \texttt{final Trust-MAPS dataset} after SMOTE resampling. \textbf{Right.} UMAP plot showing the distribution of sepsis sub-patients in the \texttt{final Trust-MAPS dataset} after SMOTE resampling.}
    \label{fig:UMAP_sepsis_nonsepsis}
\end{figure}

Next, we give the UMAP for the \texttt{final dataset with Trust-MAPS} and the \texttt{imputed dataset without Trust-MAPS} after sampling in Figures \ref{fig:UMAP1} and \ref{fig:UMAP2} respectively. Finally, we present UMAP plots after SMOTE re-sampling to overcome class imbalance, in Figure \ref{fig:UMAP_sepsis_nonsepsis}. 


\begin{figure}[t]
    \centering
    \includegraphics[scale = 0.3]{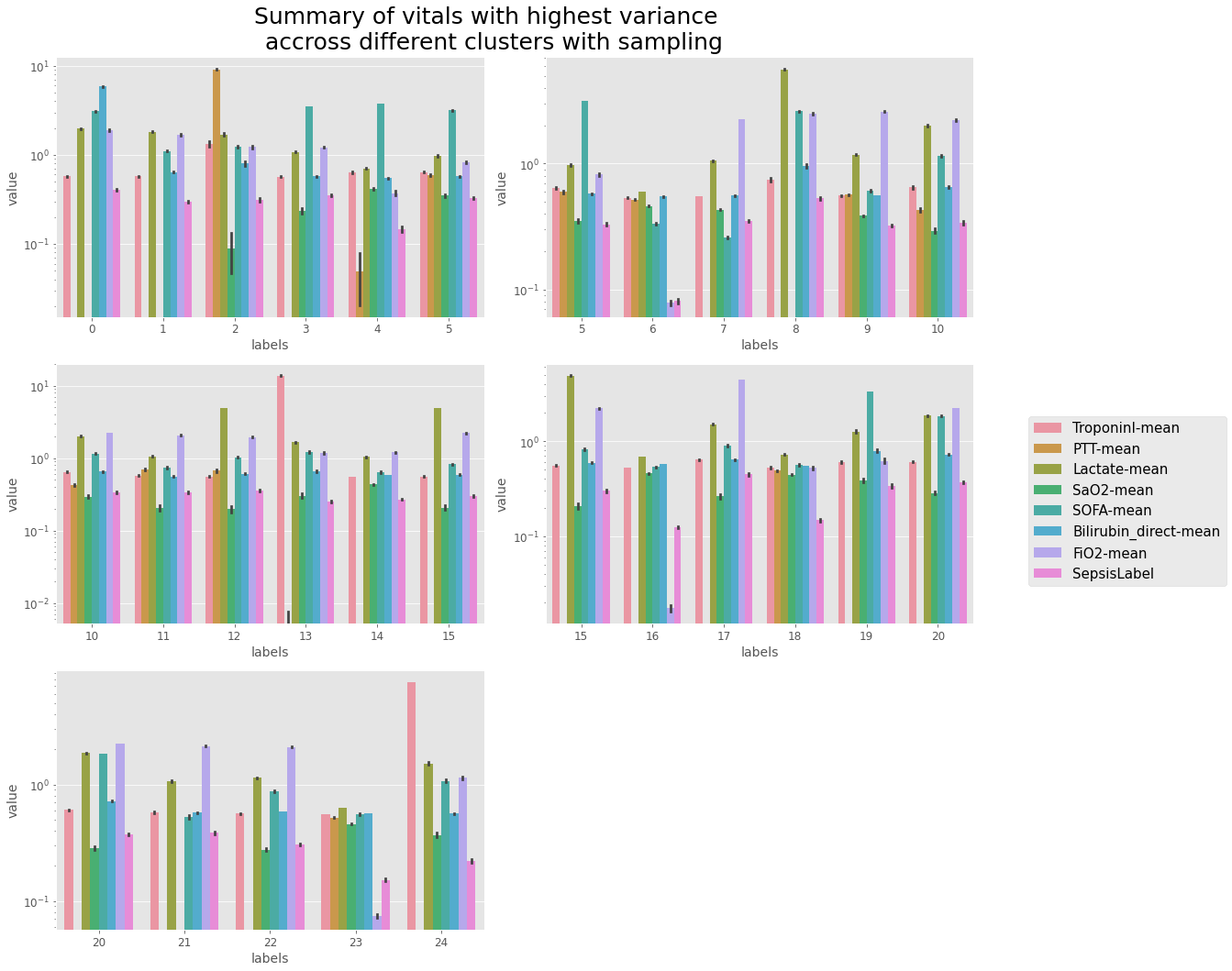}
    \caption{The seven vitals (after being averaged across the 6-hour time interval for each sub-patient) that have the highest variance across the 25 clusters for the projected data-set (\texttt{final dataset} after sampling ($\mathcal{X}^{\mathcal{F}}_S$)) appended with the sepsis concentration for each cluster.}
    \label{fig:clustering vitals}
    \vspace{-10 pt}
\end{figure}

Next, we present concentration of sepsis patients within each cluster in the \texttt{final dataset} after sampling and the variation of the optimal decision threshold chosen by maximising the $f$-score across 25 clusters. 
Lastly, we also looked at the top seven vitals that vary the most across the 25 clusters $\mathcal{X}^{\mathcal{F}}_S$. We plot these vitals in Figure \ref{fig:clustering vitals}.

\begin{figure}[htbp]
    \centering
    \includegraphics[scale = 0.5]{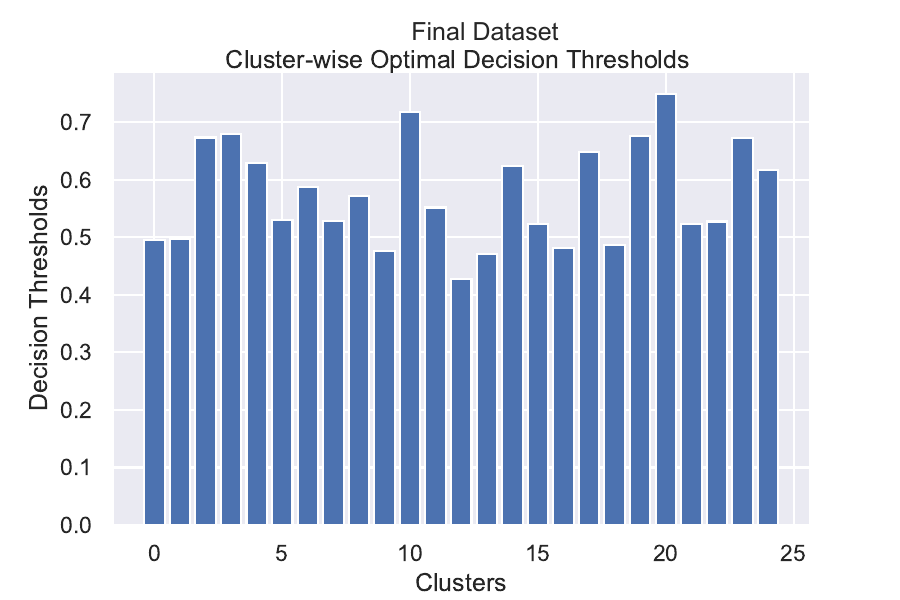}
    \includegraphics[width = 0.45\textwidth]{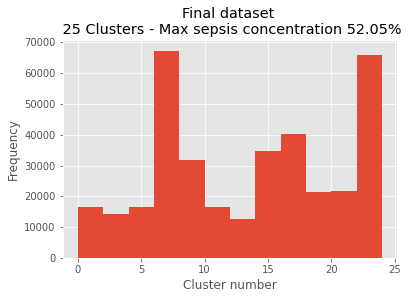}
    \caption{\textbf{Left:} Plot showing the variation of the optimal decision threshold chosen by maximising the $f$-score across 25 clusters. \textbf{Right:} Plot showing the concentration of sepsis patients within each cluster. These are results corresponding to one run of the pipeline.}
    \label{fig:thresholds}
\end{figure}

\subsection{Testing the Meta-Algorithm on Differing Lengths of Time Windows} \label{app:different_time_window_sec}
\begin{figure}[H]
\vspace{-1 pt}
    \centering
    \includegraphics[scale = 0.19]{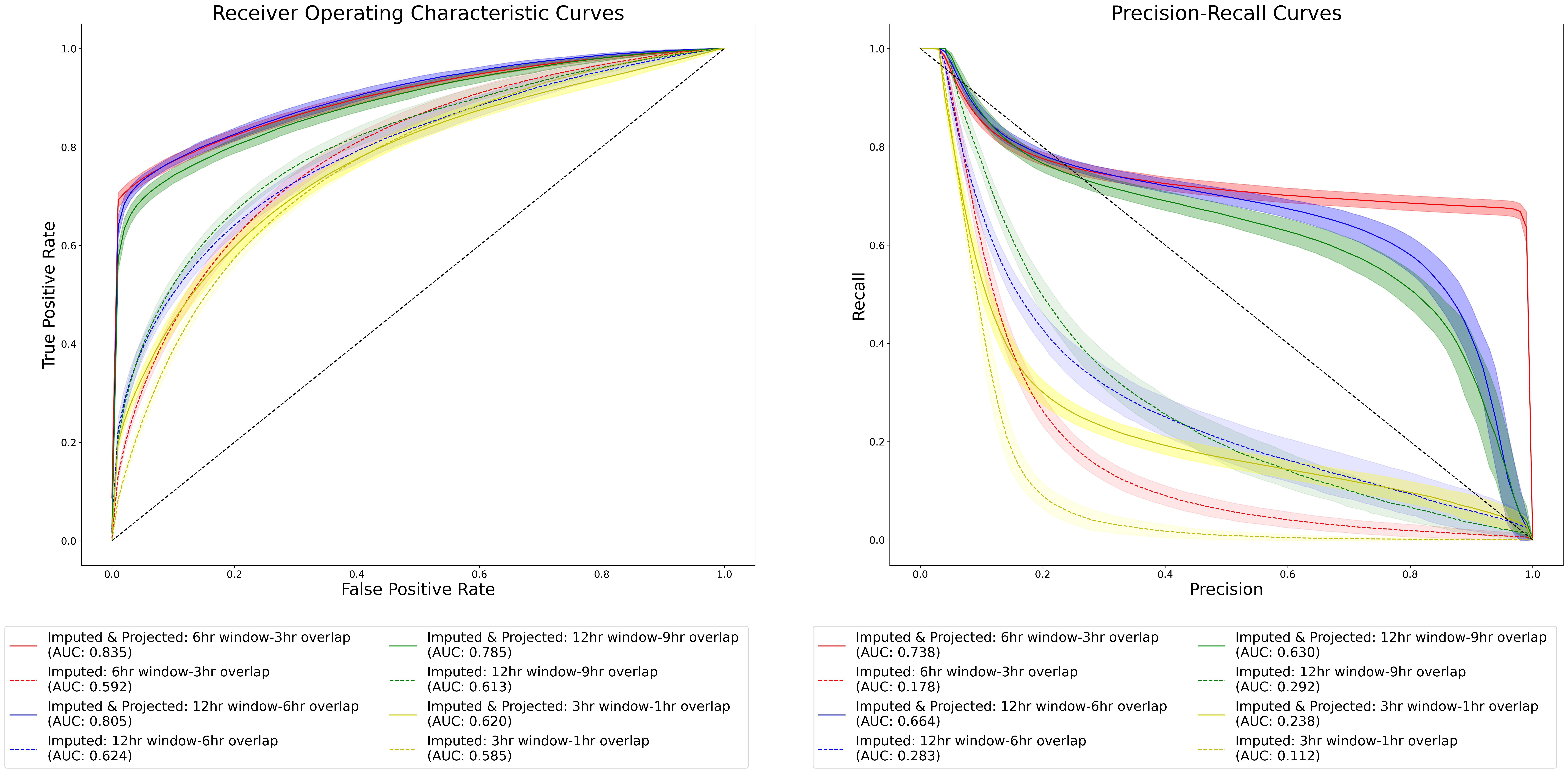}
    \caption{\textbf{Left:} A plot showing the Receiver-Operating Characteristic Curves for the results obtained on the test set with the differing time windows. \textbf{Right:} A plot showing the Precision-Recall curves on the test set with the differing time windows. We observe that a 6-hr time window with 3-hr overlap gives us the best performance in AUROC (0.835) as well as AUPRC (0.738).}
    \label{fig:different_time_windows}
\end{figure}

We tested our projections and machine learning pipeline on varying time windows and overlaps, and include the results obtained on the test set for the meta-algorithm in Table \ref{tab:different_time_windows} and Figure \ref{fig:different_time_windows}. The combination of time windows and overlaps experimented with were as follows: (i) six-hour time window with 3 hours of overlap; (ii) 12-hour time window with 6-hours of  overlap; (iii) 12-hour time window with 9 hours of overlap; (iv) 3-hour time window with 1 hour of overlap.

\begin{table}[H]
\small
\centering
\resizebox{1.1\textwidth}{!}{ \hspace{-40 pt} 
\begin{tabular}{|c|c|c|c|c|c|c|c|c|}
\hline 
  & \textbf{Method} & \textbf{Dataset}   & \textbf{Sensitivity}       & \textbf{Specificity}     
& \textbf{Precision}  & \textbf{AUC-ROC} & \textbf{AUC-PRC} & \textbf{$f$-score} \\
\hline 
\multirow{6}{1.2 cm}{6 hour window, 3 hour overlap} &
	\multirow{2}{3.2 cm}{With Trust-MAPS} & 
	\footnotesize{Train Set ($\mathcal{X}^\mathcal{F}_S$)} & 0.804 ± 0.001 & 0.999 ± 0.0002
	& 0.974 ± 0.001	& 0.961 ± 0.001	& 0.857 ± 0.001 &	0.881 ± 0.003   \\ 
	& & \footnotesize{Test Set ($\mathcal{Y}^\mathcal{F}$)} & \textbf{0.633 ± 0.018}	& \textbf{0.998 ± 0.0003} & \textbf{0.951 ± 0.004} & \textbf{0.907 ± 0.008} &	\textbf{0.707 ± 0.017} &	\textbf{0.760 ± 0.012}
 \\ \cline{2-9} & 
	\multirow{2}{3.2 cm}{Without Trust-MAPS} & \footnotesize{Train Set ($\mathcal{X}^\mathcal{I}_S$)} & 0.614 ± 0.003 &	0.951 ± 0.003&	0.538 ± 0.012&	0.828 ± 0.012&	0.199 ± 0.031&	0.552 ± 0.030  \\ 
	& &\footnotesize{Test Set ($\mathcal{Y}^\mathcal{I}$)} & 0.445 ± 0.015 &	0.949 ± 0.003&	0.499 ± 0.007 &	0.785 ± 0.019 &	0.176 ± 0.012 &	0.435 ± 0.014
 \\ \cline{2-9} & 
	\multirow{2}{3.2 cm}{Prediction using SOFA Scores}& 
	\footnotesize{Train Set ($\mathcal{X}^\mathcal{F}_S$)} &  0.168 ± 0.012&	0.896 ± 0.001&	0.054 ± 0.021&	-	& - &	0.081 ± 0.081
  \\ 
	& &\footnotesize{Test Set ($\mathcal{Y}^\mathcal{F}$)} &  0.152 ± 0.020 &	0.897 ± 0.001&	0.05 ± 0.025&	-	& -&	0.075 ± 0.074
 \\ \specialrule{.2em}{.1em}{.1em}

\multirow{6}{1.2cm}{12 hour window, six-hour overlap} &
\multirow{2}{3.2 cm}{With Trust-MAPS} & 
	\footnotesize{Train Set ($\mathcal{X}^\mathcal{F}_S$)} & 0.945 ± 0.009&	0.999 ± 0.000&	0.962 ± 0.006&	0.972 ± 0.005&	0.984 ± 0.004&	0.953 ± 0.007
   \\ 
	& & \footnotesize{Test Set ($\mathcal{Y}^\mathcal{F}$)} & 0.616 ± 0.021	&0.993 ± 0.001	&0.764 ± 0.022	&0.805 ± 0.011	&0.664 ± 0.019	&0.682 ± 0.018 \\ \cline{2-9} & 
	\multirow{2}{3.2 cm}{Without Trust-MAPS} & \footnotesize{Train Set ($\mathcal{X}^\mathcal{I}_S$)} & 0.581 ± 0.020&	0.989 ± 0.002&	0.67 ± 0.046&	0.785 ± 0.010&	0.64 ± 0.025&	0.622 ± 0.026   \\ 
	& & \footnotesize{Test Set ($\mathcal{Y}^\mathcal{I}$)} & 0.263 ± 0.023	&0.985 ± 0.002	&0.406 ± 0.047&	0.624 ± 0.012	&0.283 ± 0.032	&0.318 ± 0.026 \\ \cline{2-9} & 
	\multirow{2}{3.2 cm}{Prediction using SOFA Scores}& 
	\footnotesize{Train Set ($\mathcal{X}^\mathcal{F}_S$)} & 0.169 ± 0.012	&0.891 ± 0.024	&0.055 ± 0.021	&-&	-&	0.083 ± 0.002   \\ 
	& &\footnotesize{Test Set ($\mathcal{Y}^\mathcal{F}$)} &  0.174 ± 0.022	&0.885 ± 0.026	&0.054 ± 0.031&-&-&	0.082 ± 0.001 \\ \specialrule{.2em}{.1em}{.1em} 

\multirow{6}{1.2cm}{12 hour window, 9 hour overlap} &
\multirow{2}{3.2 cm}{With Trust-MAPS} & 
	\footnotesize{Train Set ($\mathcal{X}^\mathcal{F}_S$)} & 	0.937 ± 0.010	&0.998 ± 0.000	&0.949 ± 0.006	&0.967 ± 0.005	&0.98 ± 0.004	&0.943 ± 0.007   \\ 
	& & \footnotesize{Test Set ($\mathcal{Y}^\mathcal{F}$)} & 0.58 ± 0.022&	0.991 ± 0.001&	0.737 ± 0.023&	0.785 ± 0.011&	0.63 ± 0.021&	0.649 ± 0.018 \\ \cline{2-9} & 
	\multirow{2}{3.2 cm}{Without Trust-MAPS} & \footnotesize{Train Set ($\mathcal{X}^\mathcal{I}_S$)} & 0.603 ± 0.015&	0.992 ± 0.001&	0.782 ± 0.026&	0.798 ± 0.008&	0.695 ± 0.015&	0.68 ± 0.014   \\ 
	& &\footnotesize{Test Set ($\mathcal{Y}^\mathcal{I}$)} & 0.238 ± 0.023&	0.987 ± 0.002&	0.457 ± 0.032&	0.613 ± 0.011&	0.292 ± 0.024&	0.313 ± 0.023 \\ \cline{2-9} & 
	\multirow{2}{3.2 cm}{Prediction using SOFA Scores}& 
	\footnotesize{Train Set ($\mathcal{X}^\mathcal{F}_S$)} & 0.174 ± 0.001	&0.892 ± 0.010	&0.068 ± 0.022	&-&-&	0.098 ± 0.002   \\ 
	& &\footnotesize{Test Set ($\mathcal{Y}^\mathcal{F}$)} &  0.18 ± 0.001	&0.887 ± 0.015	&0.066 ± 0.025&-&-&	0.097 ± 0.003 \\ \specialrule{.2em}{.1em}{.1em}

\multirow{6}{1.2cm}{3 hour window, 1 hour overlap} &
\multirow{2}{3.2 cm}{With Trust-MAPS} & 
	\footnotesize{Train Set ($\mathcal{X}^\mathcal{F}_S$)} & 0.478 ± 0.014&	0.981 ± 0.002&	0.44 ± 0.027&	0.729 ± 0.007&	0.453 ± 0.014&	0.458 ± 0.016   \\ 
	& & \footnotesize{Test Set ($\mathcal{Y}^\mathcal{F}$)} & 0.261 ± 0.016	&0.979 ± 0.002&	0.283 ± 0.019&	0.62 ± 0.007&	0.238 ± 0.015&	0.271 ± 0.013 \\ \cline{2-9} & 
	\multirow{2}{3.2 cm}{Without Trust-MAPS} & \footnotesize{Train Set ($\mathcal{X}^\mathcal{I}_S$)} & 0.361 ± 0.019&	0.971 ± 0.003&	0.281 ± 0.025&	0.666 ± 0.010&	0.256 ± 0.031	&0.315 ± 0.020   \\ 
	& &\footnotesize{Test Set ($\mathcal{Y}^\mathcal{I}$)} & 0.202 ± 0.014&	0.969 ± 0.003&	0.17 ± 0.011&	0.585 ± 0.006&	0.112 ± 0.010&	0.184 ± 0.009 \\ \cline{2-9} & 
	\multirow{2}{3.2 cm}{Prediction using SOFA Scores}& 
	\footnotesize{Train Set ($\mathcal{X}^\mathcal{F}_S$)}  & 0.163 ± 0.000&	0.899 ± 0.012&	0.049 ± 0.002&	- & - &	0.075 ± 0.015   \\ 
	& &\footnotesize{Test Set ($\mathcal{Y}^\mathcal{F}$)}  &  0.152 ± 0.001&	0.897 ± 0.016&	0.044 ± 0.004& -&-&	0.068 ± 0.019 \\ \specialrule{.2em}{.1em}{.1em} 
\end{tabular}}
\\
    \caption{Confidence intervals for sensitivity, specificity, precision, AUC-ROC (Area under the Receiver Operating Characteristics Curve), AUC-PRC (Area under the Precision-Recall Curve), and $f$-score of over 50 training iterations, on the train and test set respectively.}
    \label{tab:different_time_windows}
\end{table}

We find that the six-hour time interval with 3 hours overlap configuration performs best. So, this is the configuration we work with and report in the main body of the paper. For completeness, we next focus on the six-hour time interval with 3 hours overlap configuration and the details of our results there.

\subsection{Supervised Learning Results} \label{app:res}
\begin{figure}[h]
    \centering
    \includegraphics[scale = 0.2]{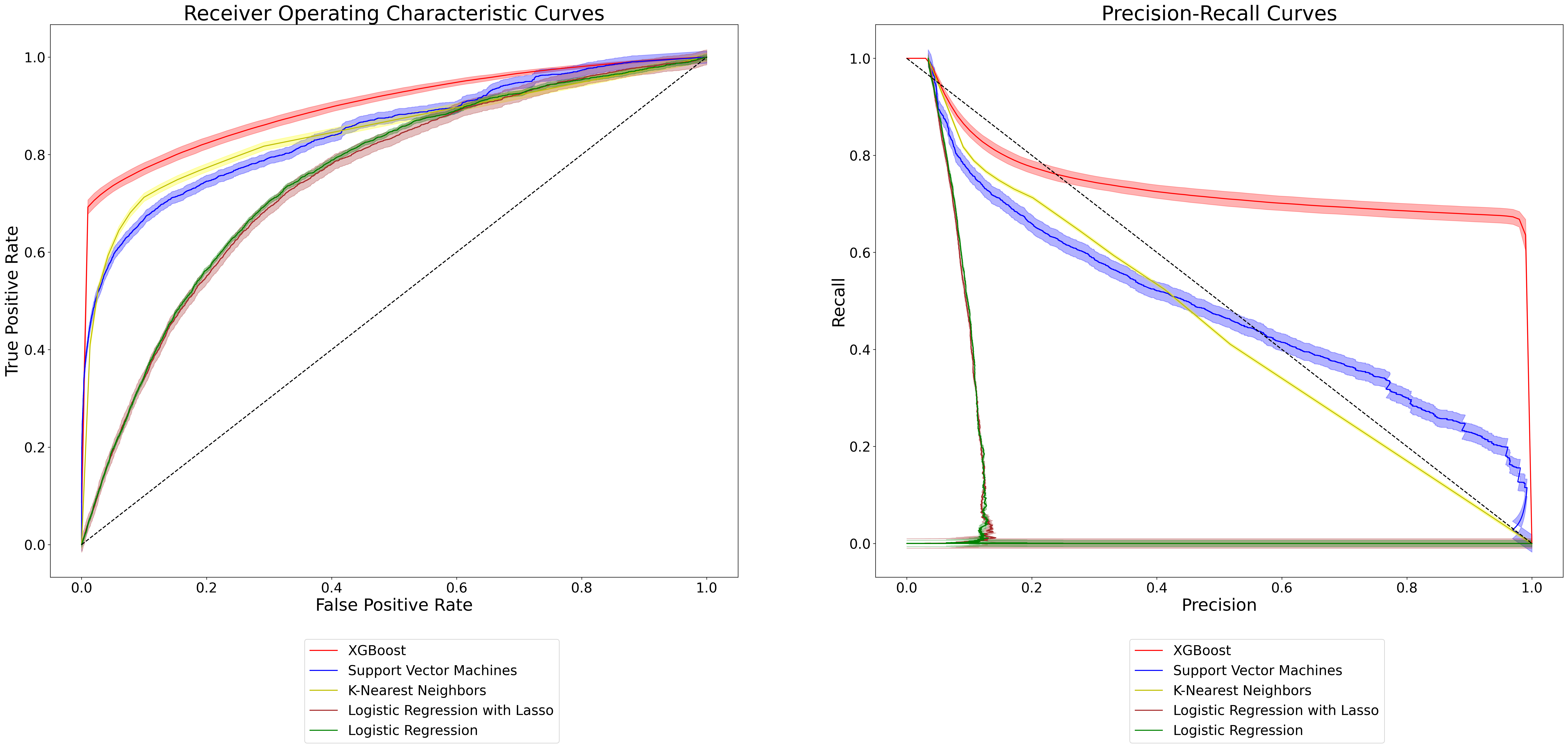}
    \caption{\textbf{Left:} A plot showing the Receiver-Operating Characteristic Curves for the results obtained on the test set with the five supervised learning algorithms discussed in our work. We observe that XGBoost outperforms the rest, with a greater area under the curve. \textbf{Right:} A plot showing the Precision-Recall curves on the test set for the five supervised learning algorithms discussed in our work. Here again, we observe that XGBoost outperforms the rest, with a greater area under the curve.}
    \label{fig:all models}
\end{figure}

In this section, we include the results obtained on the test set for each of the five supervised learning algorithms (XGBoost, SVMs, k-nearest neighbords, logistic regression with Lasso, logistic regression) that we considered in this work to predict sepsis for each of the clusters. These are listed in Table \ref{tab:all_models} and their performance is depicted in Figure \ref{fig:all models}. As is clear from the data in the table and the figure, XGBoost outperformed all other algorithms. Moreover, we also report the training time, inference time for each sub-patient, and the model size for each of those 5 models we considered in Table \ref{tab:all_models_time}, and find XGBoost to be competitive compared to other models (except logistic regression). 

\begin{table}[H]
\centering \footnotesize
\begin{tabular}{|c|c|c|c|c|c|c|}
\hline
&   \textbf{Dataset}   & \textbf{Sensitivity}       & \textbf{Specificity}     
& \textbf{Precision}  & \textbf{AU-ROC} & \textbf{$f$-score}  \\
\hline 
	\multirow{2}{4cm}{\textbf{Logistic Regression}} & 
	\footnotesize{Train Set ($\mathcal{X}^{\mathcal{F}}_S$)} & 0.460 &	0.924 &	0.176 &	0.692 &	0.254   \\ 
	& \footnotesize{Test Set ($\mathcal{Y}^{\mathcal{F}}$)} & 0.368 &	0.914 &	0.131 &	0.641 &	0.194 \\ \hline
	\multirow{2}{4cm}{\textbf{Logistic Regression with Lasso}} & \footnotesize{Train Set ($\mathcal{X}^{\mathcal{F}}_S$)} & 0.448 &	0.922 &	0.168 &	0.685 &	0.244   \\ 
	& \footnotesize{Test Set ($\mathcal{Y}^{\mathcal{F}}$)} & 0.366 &	0.913 &	0.129 &	0.640 &	0.191 \\ \hline
	\multirow{2}{4cm}{\textbf{K-nearest neighbors}} & 
	\footnotesize{Train Set ($\mathcal{X}^{\mathcal{F}}_S$)} & 0.546  &	0.999 &	0.981 &	0.718 &	0.701   \\ 
	& \footnotesize{Test Set ($\mathcal{Y}^{\mathcal{F}}$)} &  0.416 &	0.997 &	0.839 &	0.707 & 0.556 \\ \hline 
	\multirow{2}{4cm}{\textbf{Support Vector Machines}} & 
	\footnotesize{Train Set ($\mathcal{X}^{\mathcal{F}}_S$)} & 0.795 &	0.991 &	0.679 &	0.704 &	0.665   \\ 
	& \footnotesize{Test Set ($\mathcal{Y}^{\mathcal{F}}$)} & 0.696 &	0.985 &	0.630 &	0.695 &	0.661 \\ \hline
	\multirow{2}{4cm}{\textbf{XGBoost}} & 
	\footnotesize{Train Set ($\mathcal{X}^{\mathcal{F}}_S$)} & 0.804 &	0.999 &	0.974 &	0.961 &	0.881   \\ 
	& \footnotesize{Test Set ($\mathcal{Y}^{\mathcal{F}}$)} & \textbf{0.633} &	\textbf{0.998} &	\textbf{0.951} &	\textbf{0.907} &	\textbf{0.760} \\ \hline
	\multirow{2}{4cm}{\textbf{Meta-Algorithm}} & 
	\footnotesize{Train Set ($\mathcal{X}^{\mathcal{F}}_S$)} & 0.804 &	0.999 &	0.974 &	0.961 &	0.881  \\ 
	& \footnotesize{Test Set ($\mathcal{Y}^{\mathcal{F}}$)} & \textbf{0.633} & \textbf{0.998} &	\textbf{0.951} &\textbf{0.907} & \textbf{0.760} \\ 
\hline
\end{tabular}
    \caption{The average sensitivity, specificity, precision, AUC and $f$-score of all 5 supervised learning algorithms over 50 iterations, on the train and test set respectively. The meta-algorithm chose the best performing ML algorithm for each cluster based on f-score, which turned out to be XGBoost for all clusters.}
    \label{tab:all_models}
\end{table}

\begin{table}[H]
\centering
\begin{tabular}{|c|c|c|c|}
\hline
	\multirow{2}{4cm}{\centering\textbf{Model}} & 
	\multirow{1}{2cm}{\centering\textbf{Training Time (s)}} &  \multirow{1}{4cm}{\centering\textbf{Inference Time per sub-patient (ms)}} & \textbf{Model Size (Mb)} \\ & & &\\
\hline 
	\textbf{Logistic Regression} & 3.712 & 0.025 & 12 \\ \hline
	\textbf{Logistic Regression with Lasso} & 0.281 & 0.023 & 11 \\ \hline
	\textbf{K-nearest neighbors} & 14.01 & 2.071 & 353 \\ \hline
    \textbf{Support Vector Machines} & 313.4 & 2.453 & 136 \\ \hline
	\textbf{XGBoost} & 0.977 & 0.024 & 23 \\ \hline
\end{tabular}
    \caption{The average training time for each model, the average inference time for each model on one sup-patient records and the size of the model as stored in the memory.}
    \label{tab:all_models_time}
\end{table}

\subsection{Machine Learning without SMOTE} \label{app:nosmote}
There is some literature that suggests that practices such as SMOTE can bias the calibration of ML models and affect generalization \cite{10.1093/jamia/ocac093}. We have trained the sepsis prediction model without SMOTE and included the results in Table \ref{tab:smote_results}. We notice a drop in performance, with approximately a $25\%$ drop in precision. This supports our intuition that without smote, the model does not see enough samples of sepsis patient data points to learn their distribution patterns. 

\begin{table}[H]
\centering
\small\addtolength{\tabcolsep}{-1pt}
\resizebox{\textwidth}{!}{
\begin{tabular}{|c|c|c|c|c|c|c|}
\hline 
\textbf{Method}  & \textbf{Sensitivity}       & \textbf{Specificity}     
& \textbf{Precision}  & \textbf{AU-ROC} & \textbf{AU-PRC} & \textbf{$f$-score} \\
\hline 
	{With SMOTE} & 0.633 ± 0.018	& 0.998 ± 0.0003	& 0.951 ± 0.004 &	0.907 ± 0.008 &	0.707 ± 0.017 &	0.760 ± 0.012   
 \\ \hline 
	{Without SMOTE} &  0.629 ± 0.003 &	0.991 ± 0.004 &	0.682 ± 0.003 &	0.809 ± 0.005 &	0.667 ± 0.006 &	0.654 ± 0.008  
 \\ \hline
\end{tabular}}
     \caption{Sepsis Prediction Comparison for pipelines that use data preprocessing using Trust-MAPS - with and without SMOTE for training the ML model. Confidence intervals for the sensitivity, specificity, precision, AUC-ROC (Area under the Receiver Operating Characteristics Curve), AUC-PRC (Area under the Precision-Recall Curve), and $f$-score of the sepsis prediction algorithm over 50 iterations.}
    \label{tab:smote_results}
\end{table}

\subsection{Alternate Imputation Methods} \label{app:alt_imputation}

\subsubsection{ Imputation using MICE}

As an alternate imputation strategy, we use Multivariate Imputation by Chained Equations (MICE) \cite{raghunathan2001multivariate}, \cite{van2007multiple}, a commonly used imputation strategy for healthcare data. We include results of the sepsis predictions pipeline using MICE and also display the results of the imputations and subsequent projections. MICE artificially introduces values that are outside the range of observed values. Missingness at Random is an important assumption of using MICE, and results may be unreliable in cases where this assumption fails.
\vspace{10pt}

\subsubsection{ Joint Optimization of Imputation and Projection}

To explore the idea of formulating the optimization problem as a joint optimization problem for imputation as well as projection onto clinical constraints, we perform two additional experiments. The objective function was modified to suite the interpolation requirements.

Letting $V$ denote the set of clinical variables under consideration and $n$ be the length of a subpatient interval, first the EMR data $data[v, i, t]$ is imputed to get $data_{imp}[v, i, t]$. The original objective function computed the projection of a point $data_{imp}[v, i, t]$ onto the set $P$, of physiologically possible constraints. These projections are computed for each subpatient. 
\begin{equation}\label{obj1}
    \min_{x \in P} \sum_{v \in V} \sum_{i \in I} \sum_t (data_{imp}[v, i, t]  - x[v, i, t])^2 \tag{Obj$_1$}
\end{equation}

This was modified to model the joint interpolation and projection problem, and the following two objectives (Obj$_2$ and Obj$_3$) were minimized. Obj$_2$ directly computes projections, using non-missing values of original patient data. These projections are also computed for each subpatient.

\begin{equation}\label{obj2}
    \min_{x \in P} \sum_{v \in V} \sum_{i \in I} \sum_{t|data[v, i, t]\neq nan} (data[v, i, t]  - x[v, i, t])^2, \tag{Obj$_2$}
\end{equation}
 
Obj$_3$ computes projections using non-missing values using original patient data, but also penalizes high rate of change of the variables.

\begin{equation}\label{obj3}
\small{
    \min_{x \in P} \sum_{v \in V} \sum_{i \in I} \sum_{t=0|data[v, i, t]\neq nan}^{n} (data[v, i, t]  - x[v, i, t])^2 + \sum_{v \in V} \sum_{i \in I} \sum_{t=1|data[v, i, t] == nan}^{n} (x[v, i, t] - x[v, i, t-1])^2}, \tag{Obj$_3$}
\end{equation}

We find that the third objective \ref{obj3} obtains results closer to our original pipeline, with linear interpolation followed by \ref{obj1}.

\begin{figure}[h!]
    \centering
    \begin{minipage}{\textwidth}
    \begin{subfigure}[b]{0.5\textwidth}
        \centering
        \includegraphics[width=\linewidth]{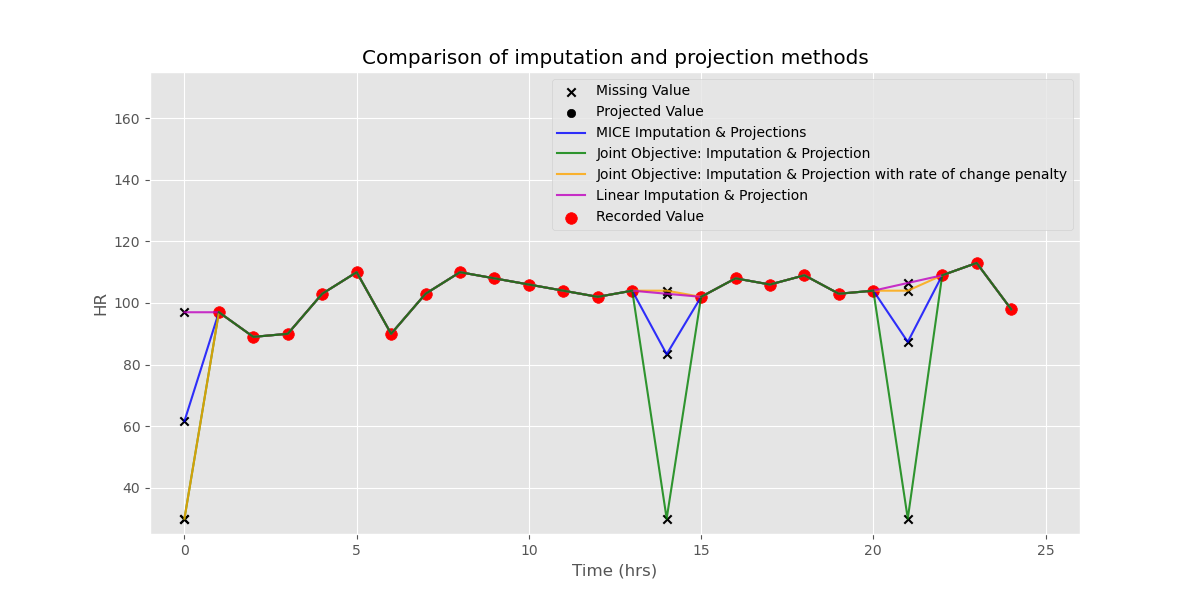}
        \caption{Heart Rate data of a single patient}
        \label{fig:imageA1}
    \end{subfigure}
    \hfill
    \begin{subfigure}[b]{0.5\textwidth}
        \centering
        \includegraphics[width=\linewidth]{Figures/Temp__Comparison_a01_fisrt.png}
        \caption{Temperature data of a single patient}
        \label{fig:imageB1}
    \end{subfigure}
     \end{minipage}

     \begin{minipage}{\textwidth}
    \begin{subfigure}[b]{0.5\textwidth}
        \centering
        \includegraphics[width=\linewidth]{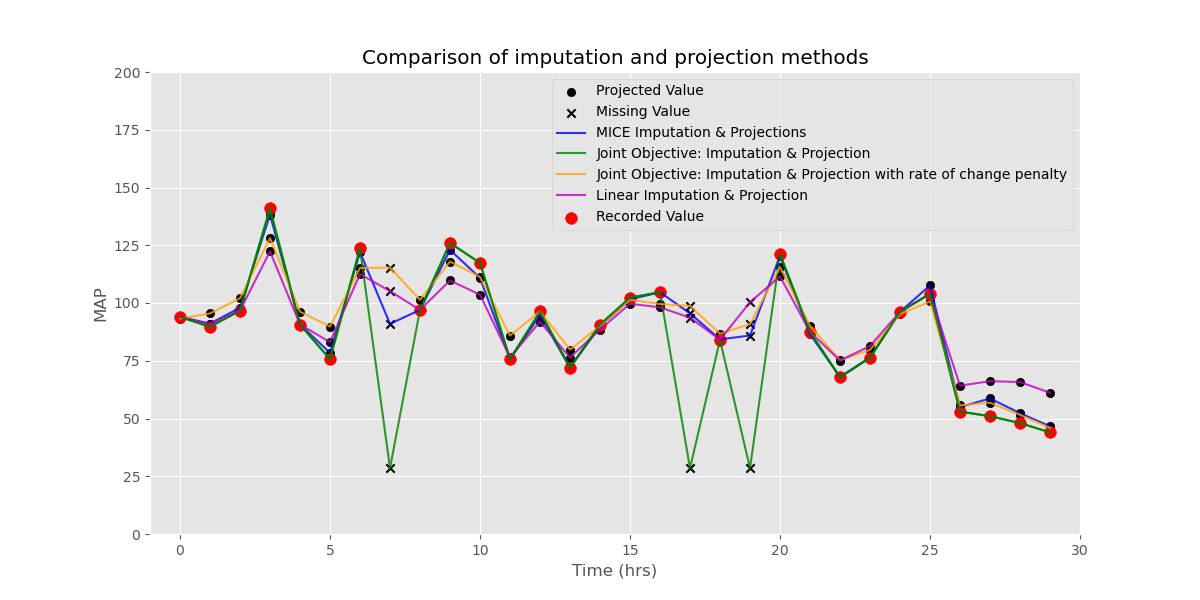}
        \caption{Mean Arterial Pressure data of a single patient}
        \label{fig:imageA2}
    \end{subfigure}
    \hfill
    \begin{subfigure}[b]{0.5\textwidth}
        \centering
        \includegraphics[width=\linewidth]{Figures/BaseExcess__Comparison_a01_fisrt.png}
        \caption{Base Excess data of a single patient}
        \label{fig:imageB2}
    \end{subfigure}
     \end{minipage}

     \begin{minipage}{\textwidth}
    \begin{subfigure}[b]{0.5\textwidth}
        \centering
        \includegraphics[width=\linewidth]{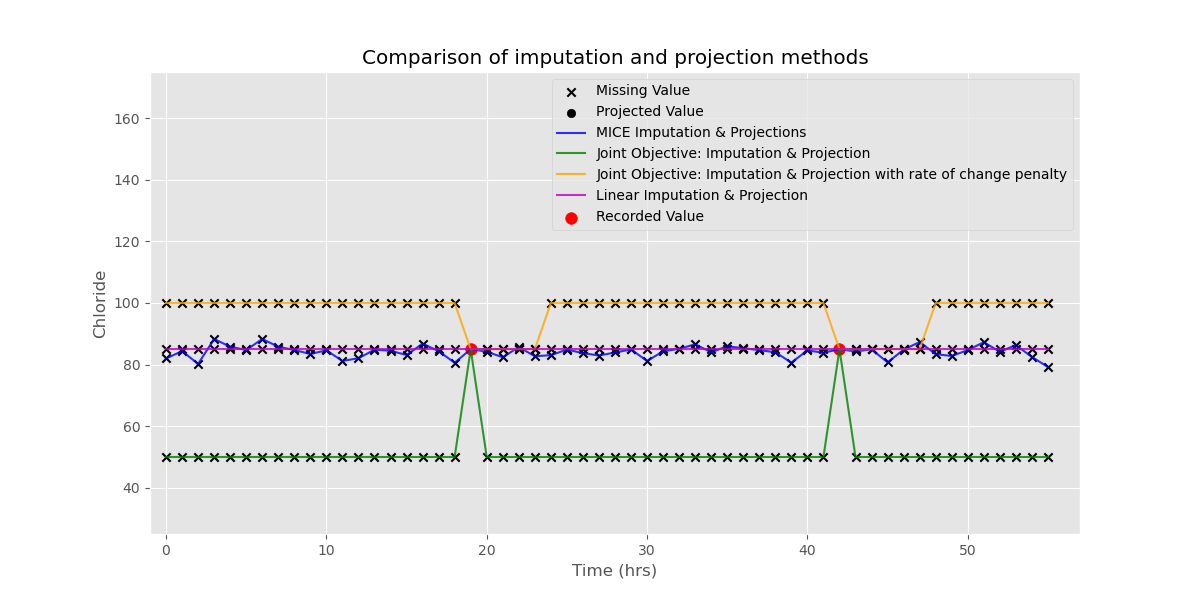}
        \caption{Chloride data of a single patient}
        \label{fig:imageA3}
    \end{subfigure}
    \hfill
    \begin{subfigure}[b]{0.5\textwidth}
        \centering
        \includegraphics[width=\linewidth]{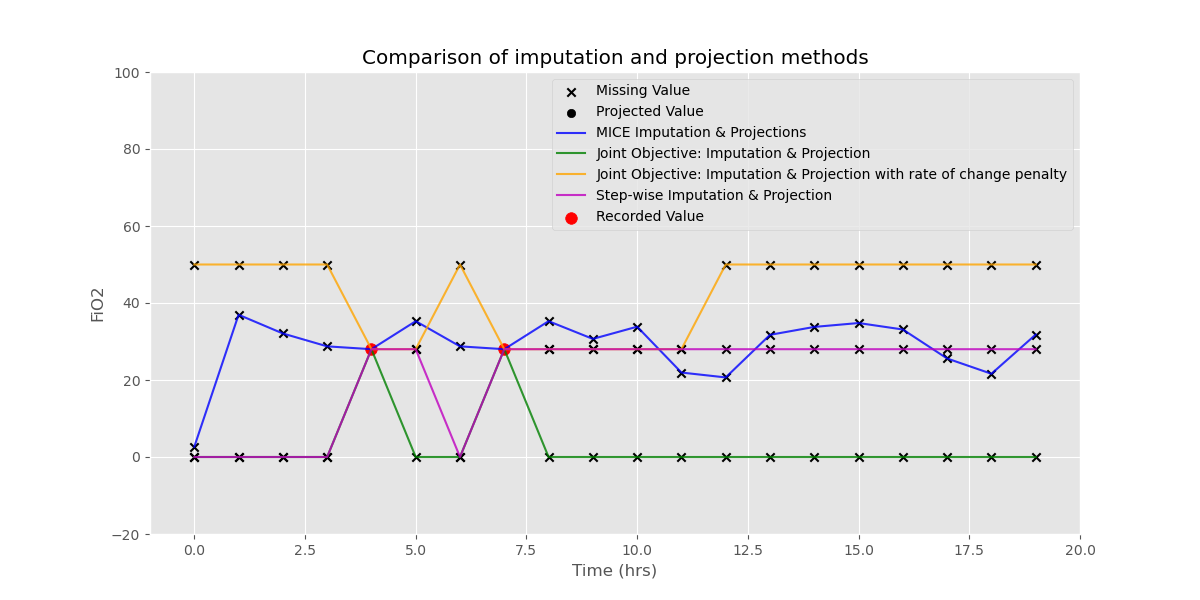}
        \caption{FiO2 data of a single patient}
        \label{fig:imageB3}
    \end{subfigure}
     \end{minipage}

    \caption{A Comparison of imputation and projection schemes for processing the data, with an example of (a) Heart Rate, (b) Temperature, (c) Mean Arterial Pressure, (d) Base Excess, (e) Chloride, (f) FiO2, of one patient from the dataset. Missing values are displayed as crosses, and recorded values are displayed as red dots. We compare the output of four processing pipelines, namely, Linear interpolation with Projections, MICE Imputation with Projections, Joint Optimization of Imputation and Projections, Joint Optimization of Imputation and Projections with rate of change constraints. }
    \label{fig:alternate_imp}
\end{figure}

Performance comparison for training the machine learning model for sepsis prediction using these alternate imputation methods is detailed in Table \ref{tab:results_impute}. A visual display of how values for different clinical variables change according to the imputation method is shown in Figure \ref{fig:alternate_imp}. The key observations were as follows:
\begin{enumerate}
    \item MICE imputation introduces values that are unseen in the data, potentially biasing and confusing the results. MICE also introduces values (Figure \ref{fig:alternate_imp})(b) that may be outside the ``normal" physiology bounds, adding to the ``normal-distance" scores that we calculate as the next step in our pipeline. This affects downstream machine learning results, as we can see a significant drop in prediction performance using MICE.
    \item Optimizing \ref{obj2} defaults missing values to the lowest possible value according to the constraints. This biases the ''distance-to-normal'' calculations and negatively impacts predictive performance.
    \item Applying objective \ref{obj3} shows better results, almost similar to linear interpolation at points where there are observed values in the neighborhood of the missing points. The drawback of this approach is poorer performance for variables that have sparsely recorded values, or no initial recorded value. Here, imputing the mean of the normal range leads to better bias mitigation in training the machine learning model.
\end{enumerate}

\begin{table}[h!]
\centering
\small\addtolength{\tabcolsep}{-1pt}
\resizebox{\textwidth}{!}{
\begin{tabular}{|c|c|c|c|c|c|c|}
\hline 
\textbf{Method}  & \textbf{Sensitivity}       & \textbf{Specificity}     
& \textbf{Precision}  & \textbf{AU-ROC} & \textbf{AU-PRC} & \textbf{$f$-score} \\
\hline 
     {Linear Imputation and Projection (Ours)} & 0.633 ± 0.018	& 0.998 ± 0.0003	& 0.951 ± 0.004 &	0.907 ± 0.008 &	0.707 ± 0.017 &	0.760 ± 0.012   
 \\ \hline 
     {MICE Imputation and Projection} &  0.370 ± 0.002 &	0.958 ± 0.001 &	0.165 ± 0.001 &	0.833 ± 0.002 &	0.248 ± 0.001 &	0.228 ± 0.012  
 \\ \hline
     {Joint Objective for Imputation and Projection} &  0.410 ± 0.001 &	0.965 ± 0.002 &	0.210 ± 0.002 &	0.854 ± 0.005 &	0.340 ± 0.008 &	0.278 ± 0.015  
      \\ \hline
     Joint Objective for Imputation and Projection\\ (with rate constraints) &  0.575 ± 0.001 &	0.997 ± 0.001 &	0.857 ± 0.010 &	0.898 ± 0.001 &	0.665 ± 0.002 &	0.688 ± 0.004  
 \\ \hline
\end{tabular}}
     \caption{Sepsis Prediction Comparison for pipelines with alternate imputation strategies in terms of the sensitivity, specificity, precision, AUC-ROC (Area under the Receiver Operating Characteristics Curve), AUC-PRC (Area under the Precision-Recall Curve), and $f$-score of the sepsis prediction algorithm.}
    \label{tab:results_impute}
\end{table}

\clearpage
\section{Model Calibration Analysis}
\label{app:calibration}

\begin{figure}[H]
    \centering
    \includegraphics[width=0.5\linewidth]{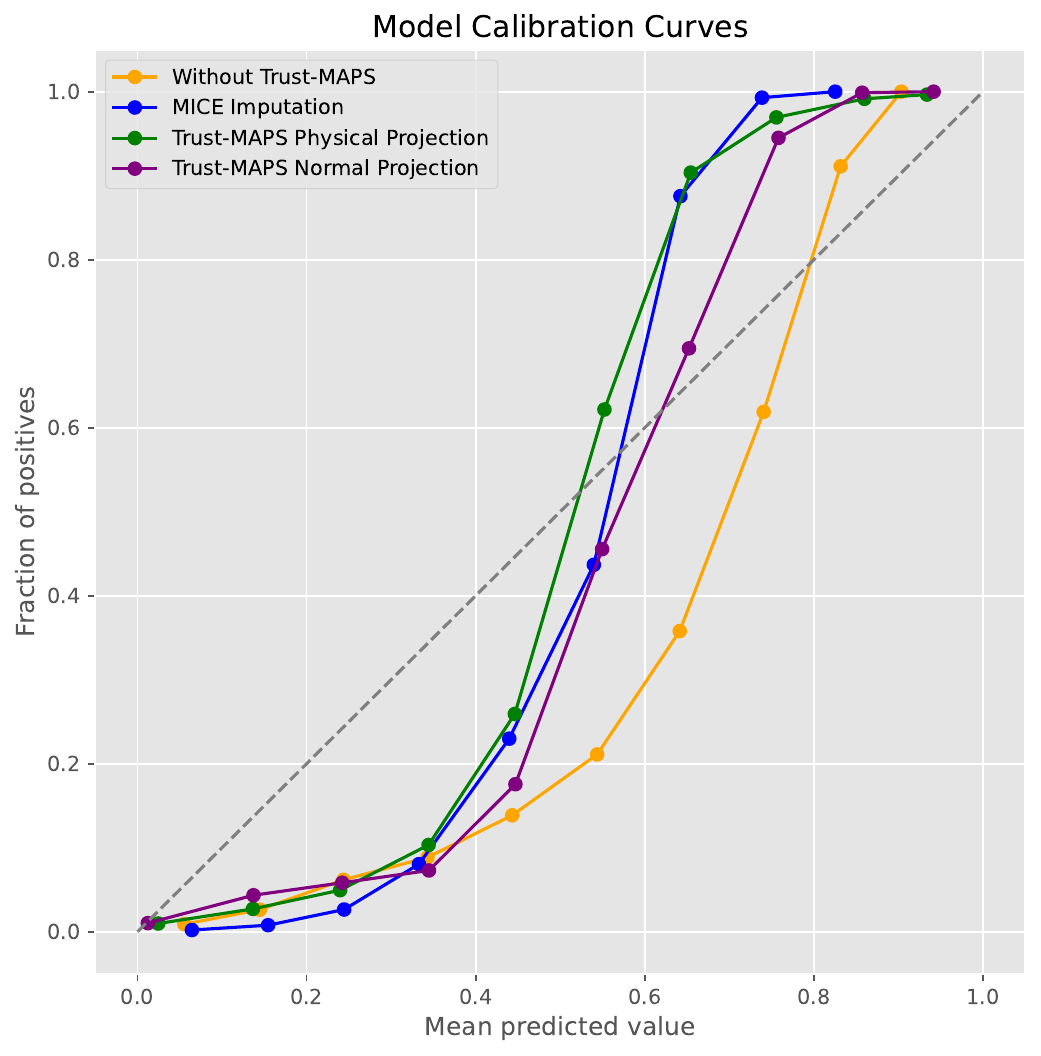}
    \caption{Reliability Curves for Sepsis Prediction models trained with and without Trust-MAPS.Lines closer to the line of the perfect model represents well-calibrated prediction probabilities.}
    \label{fig:model_calibration}
\end{figure}

We test the calibration of our trained machine learning models for the task of sepsis prediction, comparing models trained with Trust-MAPS to baseline models trained without Trust-MAPS. We use the brier score to quantify miscalibration (Table \ref{tab:brier}), and visualize results by plotting a reliability curve (Figure \ref{fig:model_calibration}).We notice that the models trained with Trust-MAPS are better calibrated.

\begin{table}[H]
\centering
\begin{tabular}{|l|l|}
\hline
Method                         & Brier Score     \\ \hline
Trust-MAPS Normal Projection   & \textbf{0.0126} \\ \hline
Trust-MAPS Physical Projection & 0.0152          \\ \hline
MICE Imputation & 0.0521          \\ \hline
Baseline (Without Trust-MAPS)             & 0.0492          \\ \hline
\end{tabular}
\caption{Brier Scores for sepsis prediction models trained with and without Trust-MAPS. A lower brier score signifies better model probability calibration} \label{tab:brier}
\end{table}

\clearpage


\end{document}